\documentclass[10pt,twocolumn,letterpaper]{article}

\usepackage[pagenumbers]{vat} 
\usepackage{multirow}
\usepackage{amsmath}
\usepackage{algorithm}
\usepackage{algpseudocode}
\usepackage{amsfonts}
\definecolor{cvprblue}{rgb}{0.21,0.49,0.74}
\usepackage[pagebackref,breaklinks,colorlinks,allcolors=cvprblue]{hyperref}

\title{
3D representation in 512-Byte: 

Variational tokenizer is the key for autoregressive 3D generation
}

\author{Jinzhi Zhang, Feng Xiong*, Mu Xu\\
AMAP, Alibaba \\
{\{wushou.zjz, xf250971, xumu.xm\}@alibaba-inc.com~}
}
\begin{document}
\maketitle
\renewcommand{\thefootnote}{}
\footnotetext{*~Equal contribution.}


\begin{abstract}
{Autoregressive} transformers have revolutionized high-fidelity image generation. One crucial ingredient lies in the tokenizer, which compresses high-resolution image patches into manageable discrete tokens {with a scanning or hierarchical order} suitable for large language models. 
Extending these tokenizers to 3D generation, however, presents a significant challenge: 
unlike image patches that naturally exhibit {spatial sequence and multi-scale relationships}, 
3D data lacks an inherent order, making it difficult to compress into fewer tokens while preserving structural details.
To address this, we introduce the Variational Tokenizer (VAT), which transforms unordered 3D data into compact latent tokens with an implicit hierarchy, suited for efficient and high-fidelity {coarse-to-fine} autoregressive modeling. VAT begins with an in-context transformer, which compress numerous unordered 3D features into a reduced token set with minimal information loss. This latent space is then mapped to a Gaussian distribution for residual quantization, with token counts progressively increasing across scales. 
In this way, tokens at different scales naturally establish the interconnections by allocating themselves into different subspaces within the same Gaussian distribution, facilitating discrete modeling of token relationships across scales.
During the decoding phase, a high-resolution triplane is utilized to convert these compact latent tokens into detailed 3D shapes. 
Extensive experiments demonstrate that VAT enables scalable and efficient 3D generation, outperforming existing methods in quality, efficiency, and generalization.
Remarkably, VAT achieves up to a 250$\times$ compression, reducing a 1MB mesh to just 3.9KB with a 96\% F-score, and can further compress to 256 int8 tokens, achieving a 2000$\times$ reduction while maintaining a 92\% F-score.

\end{abstract}

\section{Introduction}
\label{sec:intro}

A growing trend in 3D generation is the shift from traditional image-based methods to 3D native generation modeling. Conventional approaches, such as Large Reconstruction Models~(LRMs)~\cite{DBLP:conf/iclr/Hong0GBZLLSB024, DBLP:journals/corr/abs-2403-02151, DBLP:journals/corr/abs-2404-12385, DBLP:conf/eccv/TangCCWZL24} and Score Distillation Sampling~(SDS)~\cite{DBLP:conf/cvpr/WangDLYS23, DBLP:conf/iclr/PooleJBM23, DBLP:conf/nips/Wang00BL0023}, rely heavily on multi-view image inputs, making them highly sensitive to image quality and often resulting in low-fidelity 3D models. Recently, 3D native generation methods~\cite{DBLP:journals/tog/ZhangTNW23, DBLP:journals/corr/abs-2405-14832, DBLP:conf/nips/ZhaoLCZWCFCYG23, DBLP:journals/corr/abs-2405-14979, DBLP:conf/eccv/LanHYZMDPL24, DBLP:journals/corr/abs-2305-02463, DBLP:journals/corr/abs-2409-12957} have employed diffusion models in 3D latent spaces using 3D variational auto-encoders~(VAEs)~\cite{DBLP:journals/corr/KingmaW13}. However, these approaches face significant challenges in scalability and require lengthy training times, limiting their practical applicability.

In parallel, AutoRegressive~(AR) based Large Language Models~(LLMs)~\cite{radford2019language} have ushered in a new era in artificial intelligence. These models have revolutionized high-fidelity image and video generation~\cite{DBLP:journals/corr/abs-2406-06525, DBLP:conf/iclr/YuLGVSMCGGHG0ER24, DBLP:conf/icml/KondratyukYGLHS24, DBLP:journals/corr/abs-2404-02905}, demonstrating exceptional scalability, generality, and versatility. A crucial component of these models is the tokenizer, which compresses input data into discrete tokens, enabling AR models to leverage self-supervised learning for next-token or next-scale prediction.

However, extending these models to 3D tasks poses significant challenges, primarily due to the difficulty of efficiently compressing unordered 3D features. Unlike images, which can be easily tokenized into 2D grids while preserving spatial relationships and hierarchical structures, 3D data lacks inherent spatial continuity.
For example, current attempts to reformulate unordered 3D features into 2D triplanes~\cite{wu2024direct3d} or 1D latents~\cite{DBLP:journals/tog/ZhangTNW23} struggle to learn effective token sequences from these compressed latent space. 
Similarly, methods such as MeshGPT~\cite{DBLP:conf/cvpr/SiddiquiAATSRDN24} tokenize serialized mesh data using a GNN-based encoder~\cite{Zhou2018GraphNN}. However, these approaches rely on manually defined sequences on unordered graphs~\cite{DBLP:conf/cvpr/00020WLL0O0Z24}, which limits their ability to generalize to complex datasets.
Instead of imposing an artificial order on 3D data, G3PT~\cite{DBLP:journals/corr/abs-2409-06322} proposes scalable AR modeling using next-scale rather than next-token prediction by mapping 3D data into coarse-to-fine 1D latent tokens.
However, the latent 1D token space lacks meaningful semantic representation at coarse levels. Unlike images, which naturally benefit from pyramid-like hierarchical features, G3PT struggles to compress 3D features into a compact token set without sacrificing the level-of-detail hierarchy, thereby limiting its ability to generate high-fidelity meshes.

\textit{Why do AR models in 3D lag behind their counterparts in visual generation?} This paper argues that a key factor is the absence of an effective tokenizer capable of compressing complex 3D features into a set of latent distributions while preserving their interconnections.
Our core idea is straightforward: the 3D input features are first compacted into a Gaussian distribution, and multi-scale token maps are then allocated to its subspaces.
In this way, by starting from a single token map and progressively predicting higher-scale token maps conditioned on previous ones, next-scale AR modeling easily learns the multi-scale sequential relationships inherent in different subspaces.

\begin{figure}[t]
\centering
\includegraphics[width=0.45\textwidth]{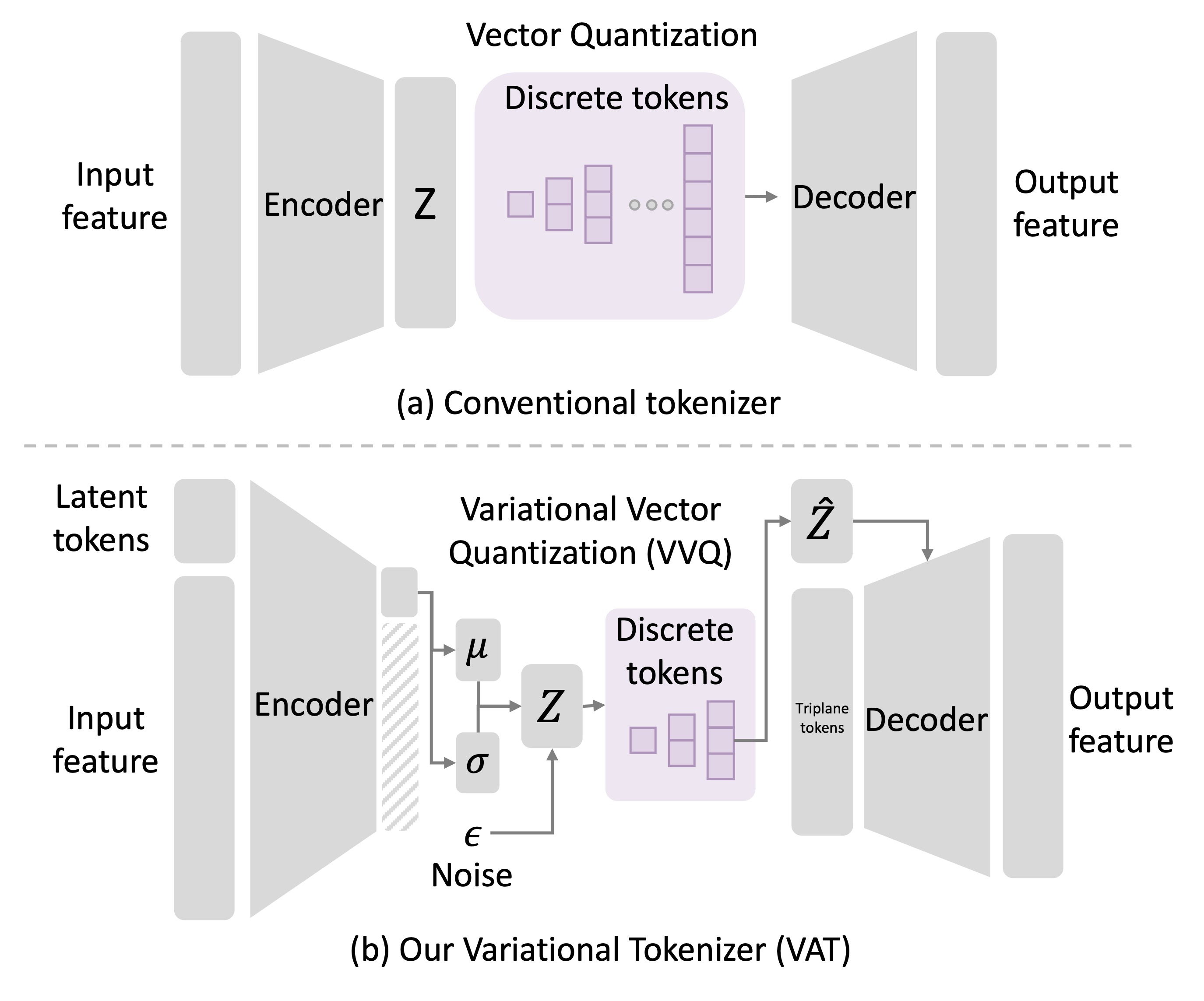} 
\caption{Comparison between (a) conventional tokenizer and (b) our proposed Variational Tokenizer (VAT). In (a), an encoder transforms input features into latent embeddings $Z$, which are directly quantized into discrete tokens. In (b), VAT employs an in-context transformer to compress unordered input features into a reduced token set, which is then mapped to a Gaussian distribution. 
Quantization is residually applied across scales, allowing tokens to self-organize into distinct subspaces within the same Gaussian distribution, enabling autoregressive next-scale token prediction. 
}
\label{fig: method_vq}
\end{figure}

To this end, we propose the Variational Tokenizer (VAT), which comprises a transformer encoder, a Variational Vector Quantizer (VVQ), and a triplane decoder. 
%
As shown in Fig.~\ref{fig: method_vq}, during tokenization, the 3D input features are concatenated with a smaller 1D sequence of latent tokens and processed by a transformer encoder. The encoder's output retains only the latent tokens, resulting in a compact 1D latent representation that preserves the original information.
Next, VVQ maps the 1D latent onto a Gaussian distribution, where quantization is applied residually across scales. This process allows tokens to self-organize into distinct subspaces within the same Gaussian distribution.
%
%
%
Following vector quantization, the triplane decoder recovers the output features based on the discrete token maps, and a triplane-based convolutional neural network, combined with an MLP, upsamples the low-resolution features into a high-resolution 3D occupancy grid.

We empirically demonstrate that VAT enables scalable and efficient 3D generation, outperforming existing methods in both quality and generalization.
%
More impressively, as shown in Fig.~\ref{fig: method_teaser}, VAT achieves a 250-fold compression, reducing an 1MB mesh to just 3.9KB with a 96\% F-score, and can further compress data to 256 int8 tokens with a codebook size of 256, resulting in a 2000-fold reduction while maintaining a 94\% F-score.

\section{Related Work}
\subsection{Native 3D Generation}
With advances in neural 3D representations~\cite{DBLP:conf/cvpr/ChenZ19, DBLP:conf/eccv/MildenhallSTBRN20, DBLP:conf/iclr/CardaceRBZSS24} and the availability of large-scale 3D datasets~\cite{DBLP:conf/cvpr/DeitkeSSWMVSEKF23, DBLP:conf/nips/DeitkeLWNMKFLVG23}, researchers have increasingly focused on high-fidelity native 3D generation, falling into two main categories: Diffusion-based and Auto-regressive (AR)-based approaches. 
Several works~\cite{DBLP:journals/tog/ZhangTNW23, DBLP:journals/corr/abs-2405-14832, DBLP:conf/nips/ZhaoLCZWCFCYG23, DBLP:journals/corr/abs-2405-14979, DBLP:conf/eccv/LanHYZMDPL24, DBLP:journals/corr/abs-2305-02463, DBLP:journals/corr/abs-2409-12957} use a VAE~\cite{DBLP:journals/corr/KingmaW13} to compress 3D data into a compact latent format, simplifying training for latent diffusion models. Notably, CLAY~\cite{DBLP:journals/tog/ZhangWZQPJYXY24} 
scales to large datasets and generalize effectively across diverse input conditions. 
Other approaches~\cite{DBLP:conf/cvpr/SiddiquiAATSRDN24, DBLP:journals/corr/abs-2406-10163, DBLP:journals/corr/abs-2408-02555, DBLP:journals/corr/abs-2409-18114} use face sorting to tokenize 3D meshes, compressing them with VQ-VAE~\cite{DBLP:conf/nips/OordVK17} and generating sequences via an auto-regressive transformer. However, these methods struggle with the unordered nature of 3D data, limiting their generalization.

A recent advancement, G3PT~\cite{DBLP:journals/corr/abs-2409-06322}, employs cross-scale vector quantization to implement 3D multi-scale VQ-VAE, using a next-scale AR approach to generate 3D geometry from coarse to fine
Building on this, we adopt the next-scale AR approach and introduce a stochastic VQ-VAE and Triplane Decoder for more sophisticated 3D geometry generation.


\subsection{Token Compression}
Token compression reduces computational load by minimizing the number of tokens while retaining essential information. Some methods~\cite{DBLP:conf/nips/RaoZLLZH21, DBLP:conf/ijcai/LiuWG23, DBLP:conf/iclr/BolyaFDZFH23} dynamically prune non-essential tokens through filtering or merging. Llama-VID~\cite{DBLP:conf/eccv/LiWJ24} uses average pooling with a learnable linear layer, while MiniCPM-VL~\cite{DBLP:journals/corr/abs-2408-01800} employs cross-attention with a fixed number of queries. However, these methods lose valuable visual information at higher compression rates. TiTok~\cite{DBLP:journals/corr/abs-2406-07550} combines visual tokens with a 1D sequence of latent tokens, using self-attention for in-context compression, significantly reducing information loss. 

\section{Method}

We present the Variational Tokenizer (VAT), which facilitates efficient and high-fidelity 3D generation through next-scale autoregressive modeling. The 3D generation process consists of two stages. 
In the first stage, VAT transforms unordered 3D data into coarse-to-fine compact latent tokens with an inherent hierarchy (Sec.~\ref{ch:vq}). This process starts with an in-context transformer that compresses 3D features into a compact token set, which is subsequently mapped to a Gaussian distribution, establishing structured token relationships across scales. A high-resolution triplane reconstructs these latent tokens into detailed 3D occupancy grids.
%
%
In the second stage, the autoregressive transformer leverages these multi-scale tokens by starting with a single token and progressively predicting higher-resolution 3D token maps. Each scale is conditioned on all previous scales, as well as the image or text conditions (Sec.~\ref{ch: AR_modeling}).

\subsection{Preliminary: Autoregressive Modeling}\label{ch: Preliminary}

Autoregressive modeling is widely used for generating and reconstructing 2D or 3D content through a two-stage process. In the first stage, a tokenizer compresses input $I$ into discrete tokens. The encoder maps $I$ to latent embeddings $Z$, where: $Z = \text{Enc}(I), \quad Z \in \mathbb{R}^{L \times D}$.
Then, each token $z_i$ is quantized by mapping to the nearest code $c_k$ from a codebook $C$:
\begin{equation}
x_i = \text{Quant}(z_i) = c_k, \quad k = \underset{j}{\arg\min} \| z_i - c_j \|_2.
\end{equation}
In the second stage, a causal transformer predicts these tokens via next-token prediction~\cite{wang2024emu3,cheng2023sdfusion}.

To address the lack of sequential order in 2D and 3D data, models like VAR~\cite{VAR} and CAR~\cite{g3pt} adopt next-scale prediction. The latent embeddings $Z$ is progressively quantized into different token maps $x^{(s)}$ across scales, and the token generation across scales follows the probability distribution of:
$P(x) = \prod_{s=1}^S P(x^{(s)} \mid x^{(1)}, \dots, x^{(s-1)}).
$

\subsection{Variational Tokenizer}\label{ch:vq}

As illustrated in Fig.~\ref{fig: method_vq}, we present our primary contribution: Variational Tokenizer (VAT). This method consists of a transformer encoder for in-context token compression, a Variational Vector Quantizer (VVQ) to get cross-scale discrete tokens, and a decoder for de-tokenization. 
Refer to Algo.~\ref{algo: vvq} for a detailed illustration of the algorithm.
%

\begin{figure*}[h]
\centering
\includegraphics[width=1.0\textwidth]{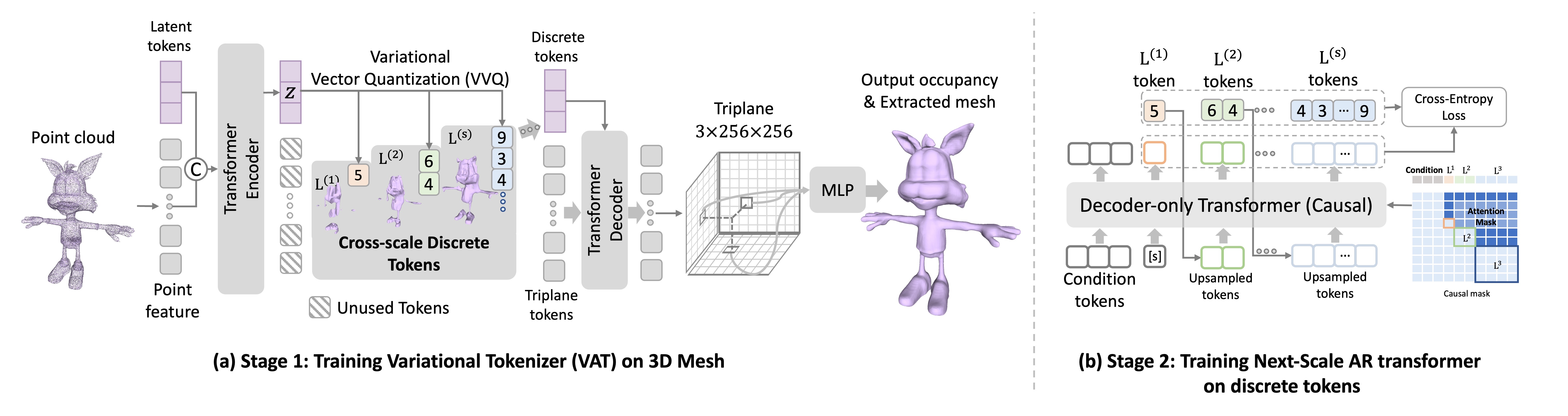} %
\caption{Overview of the two-stage training pipeline. (a) Stage 1: Training the Variational Tokenizer (VAT). The process begins with a 3D point cloud that is transformed into point features and compressed into latent tokens using a transformer encoder (Sec.~\ref{ch:vq}). Variational Vector Quantization (VVQ) maps these latent tokens onto cross-scale discrete tokens. These discrete tokens are decoded into a triplane representation, which is subsequently upsampled and processed by an MLP to generate the dense occupancy volume 
.
(b) Stage 2: Training the Next-Scale Autoregressive Transformer on discrete tokens. Here, discrete tokens generated by VAT are used as supervised signal for a decoder-only transformer trained for next-scale prediction. The model is conditioned on 
image and text features with a causal attention mask trained by cross-entropy loss (Sec.\ref{ch: AR_modeling}).
}
\label{fig: method_pipeline}
\end{figure*}

\textbf{In-context token compression.}
The tokenization process begins with an input feature \( I \in \mathbb{R}^{N \times D} \), which, in our case, represents the 3D point cloud feature. Following the 3DShape2VecSet~\cite{DBLP:journals/tog/ZhangTNW23}, we transform the point clouds \( \mathbf{P} \in \mathbb{R}^{N_p \times (3+3)} \)—consisting of positions and normals sampled from 3D object surfaces—into this feature \( I \).
More details can be found in the appendix. 
%

Subsequently, we employ an in-context token compression module to transform the feature \( I \) into an 1D sequence of latent tokens. This module achieves a high compression ratio with minimal information loss, even as the number of tokens is significantly reduced~\cite{DBLP:conf/eccv/LiWJ24}.
Specifically, the input feature $I$ is concatenated with $K$ learnable latent tokens, ${q} \in \mathbb{R}^{K \times D}$, and passed through a transformer-based encoder. Only $K$ latent tokens are retained
, producing a compact sequence of latent tokens ${Z\in \mathbb{R}^{K \times D}}$ as output. Note that $K$ is much smaller than $N$. 

\textbf{Variational Vector Quantization (VVQ).} 
%
While residual vector quantization (VQ)~\cite{DBLP:conf/nips/OordVK17} has been widely adopted in previous AR models~\cite{lee2022autoregressive,VAR}, its deterministic nature limits the tokenizer’s ability to capture inter-code correlations. This limitation becomes more evident during significant compression of the latent token space, where coarse-level tokens lose semantic richness and fail to effectively represent the underlying meaning.
To address this, we first map the encoder output onto a Gaussian distribution, then project token maps at different scales onto subspaces of this distribution. 
As a result, each token map is modeled as a Gaussian distribution, and the token maps corresponding to different subspaces are tightly linked together.



As shown in Fig.~\ref{fig: method_vq}, we first map the encoder output \( Z \) onto a Gaussian distribution characterized by mean \(\mu \in \mathbb{R}^{K \times d}\) and variance \(\sigma \in \mathbb{R}^{K \times d}\) using a linear layer. The Gaussian distribution is represented as:
    $Z_{0} = \mu + \sigma \cdot \epsilon$,
where \(\epsilon\) is sampled from a standard normal distribution \(\mathcal{N}(0, I)\).
As shown in Fig.~\ref{fig: method_vq}(a) and Fig.~\ref{fig: method_pipeline}(a), this Gaussian distribution is progressively quantized into discrete latent tokens \(x^{(s)} \in \mathbb{R}^{L^{(s)} \times D}\), where \(L^{(s)}\) denotes the number of tokens at scale \(s\). 
The quantization process at each scale is defined as:
\begin{equation}
x^{(s)} = \text{Quant}(\text{Down}(Z_s)),
\end{equation}
where \(\text{Down}(\cdot)\) represents the downsampling operation~\cite{VAR,g3pt}, projecting the Gaussian distribution into subspaces for different scales.
Starting from \(Z_0\), the residual for the next scale is updated iteratively:
\begin{equation}
Z_{s+1} = Z_s - \text{Up}(x^{(s)}),
\end{equation}
where \(\text{Up}(\cdot)\) denotes the upsampling operation~\cite{VAR,g3pt}, which project back to the same space of the input latent token feature.

Finally, the dequantized output \(\hat{Z}\) is obtained by summing the upsampled features across all scales:
\begin{equation}
\hat{Z} = \sum_{s=1}^S \text{Up}(x^{(s)}). \label{eq: dequantized}
\end{equation}


\begin{algorithm}
\caption{Variational Vector Quantization in VAT.}\label{algo: vvq}
\begin{algorithmic}[1]
\Require Raw input feature $I$
\State Initialize $(\mu, \sigma) = Z  = \text{Enc}(I \oplus {q})$, token list $X = [\;]$
\State Sample $\epsilon$ from standard Gaussian distribution $\mathcal{N}(0, I)$
\State Set initial latent $Z_{0} = \mu + \sigma \cdot \epsilon$
\For{$s = 0, \ldots, S-1$} \Comment{Iterate across scales}
    \State $x^{(s)} = \text{Quant}(\text{Down}(Z_{s}))$ \Comment{Vector quantization}
    \State Append $x^{(s)}$ to $X$ 
    \State Update residual: $Z_{s+1} = Z_s - \text{Up}(x^{(s)})$
\EndFor
\State Compute de-quantized tokens: $\hat{Z} = \sum_{s=1}^S \text{Up}(x^{(s)})$
\State \Return $X$, $\hat{Z}$
\end{algorithmic}
\end{algorithm}

\textbf{Triplane decoder.} 
To recover the content feature from $\hat{Z}$, we utilize a set of learnable tokens ${M} \in \mathbb{R}^{L \times D}$, which are spatially replicated to match the desired resolution of the output feature. These tokens  form the input to a transformer-based decoder conditioned on the quantized latent tokens $\hat{Z}$ in Eq.~\ref{eq: dequantized} using a cross-attention layer and several self-attention layers. The output feature is $\hat{\mathbf{I}}\in \mathbb{R}^{L \times D}$.


\begin{figure*}[ht]
\centering
\newcommand{\figw}{0.8}
\includegraphics[width=\figw\textwidth, keepaspectratio]{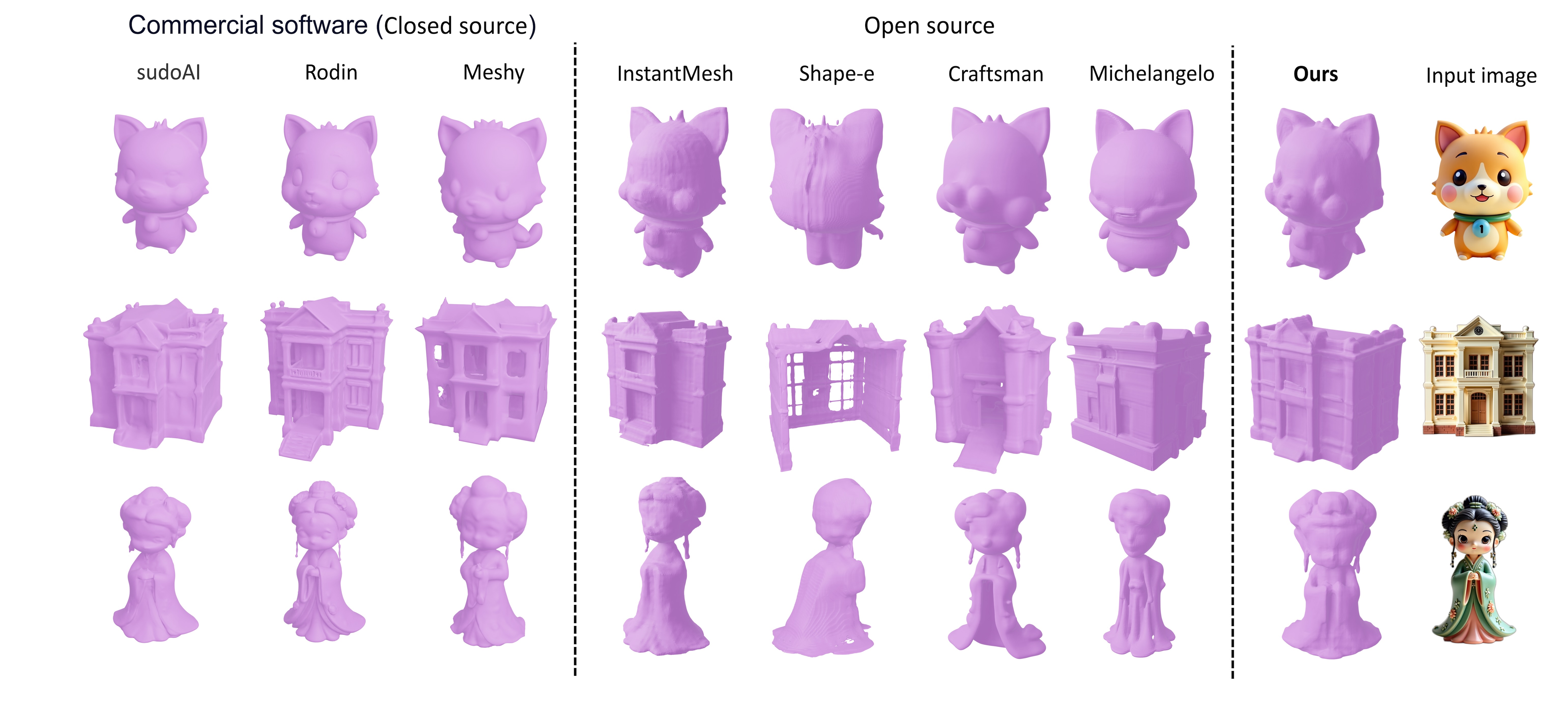} 
\caption{Comparision of state-of-the art 3D generation methods using in-the-wild images. Note that the commercial software displayed on the left may expand thousands of their own data for training, whereas our model is only trained on the Objaverse dataset.}
\label{fig: wild_compare}
\end{figure*}

As shown in Fig.~\ref{fig: method_pipeline}(a), an explicit triplane latent representation is employed to convert the latent feature ${\hat{I}}$ into 3D geometry~\cite{wu2024direct3d, wang2022rodingenerativemodelsculpting}. This process reshapes ${\hat{I}}$ into three 2D planes, yielding ${I_{tri}} \in \mathbb{R}^{3 \times r \times r \times D}$. Convolutional layers then progressively upsample ${I_{tri}}$, generating high-resolution triplane features, denoted as $\mathbf{T} = (\mathbf{T}_{XY}, \mathbf{T}_{YZ}, \mathbf{T}_{XZ})$. This approach efficiently captures intricate 3D spatial details.
%
However, direct triplane upsampling can cause blurring and aliasing artifacts at high resolutions due to insufficient sampling detail. Therefore, each triplane is represented by three mipmaps at progressively higher resolutions~\cite{barron2021mipnerf}, enabling smoother interpolation of occupancy values through an MLP-based mapping network.

%
To enhance training stability, a semi-continuous approach is used to smooth gradients near the surface, assigning binary occupancy values outside a threshold distance and continuous values within it, based on the Signed Distance Function (SDF) of each query point~\cite{wu2024direct3d}.


\subsection{Next-scale AR modeling with conditions}\label{ch: AR_modeling}




After training VAT, we obtain a set of discrete tokens, which serve as input for training the AR model. The overall framework is shown in Fig.~\ref{fig: method_pipeline}(b).
%
We use pre-trained DINO-v2 (ViT-L/14)~\cite{oquab2023dinov2} as conditional image tokens. A linear layer projects these $N_I$ image tokens \(I_{dino} \in \mathbb{R}^{L_I \times C_I}\) to match the channel dimensions of the AR model, a decoder-only transformer similar to GPT-2~\cite{radford2019language}. These image tokens are then concatenated with the cross-scale latent tokens obtained from VAT.
The start token $[s]$ serves as a text condition, obtained by extracting a text prompt from a pre-trained CLIP model~\cite{radford2021learning} (ViT-L/14). 

The AR process begins with a single token map and progressively predicts higher-scale token maps conditioned on previous ones. At each scale \(s\), all tokens at scale \(L^{(s)}\) are generated in parallel, conditioned on previous tokens and their positional embeddings. During training, a block-wise causal attention mask ensures that each token map at \(L^{(s)}\) can only attend to its prefix. During inference, kv-caching~\cite{pope2023efficiently} is employed for efficient sampling.

\begin{figure*}[t]
\centering
\includegraphics[width=0.9\textwidth]{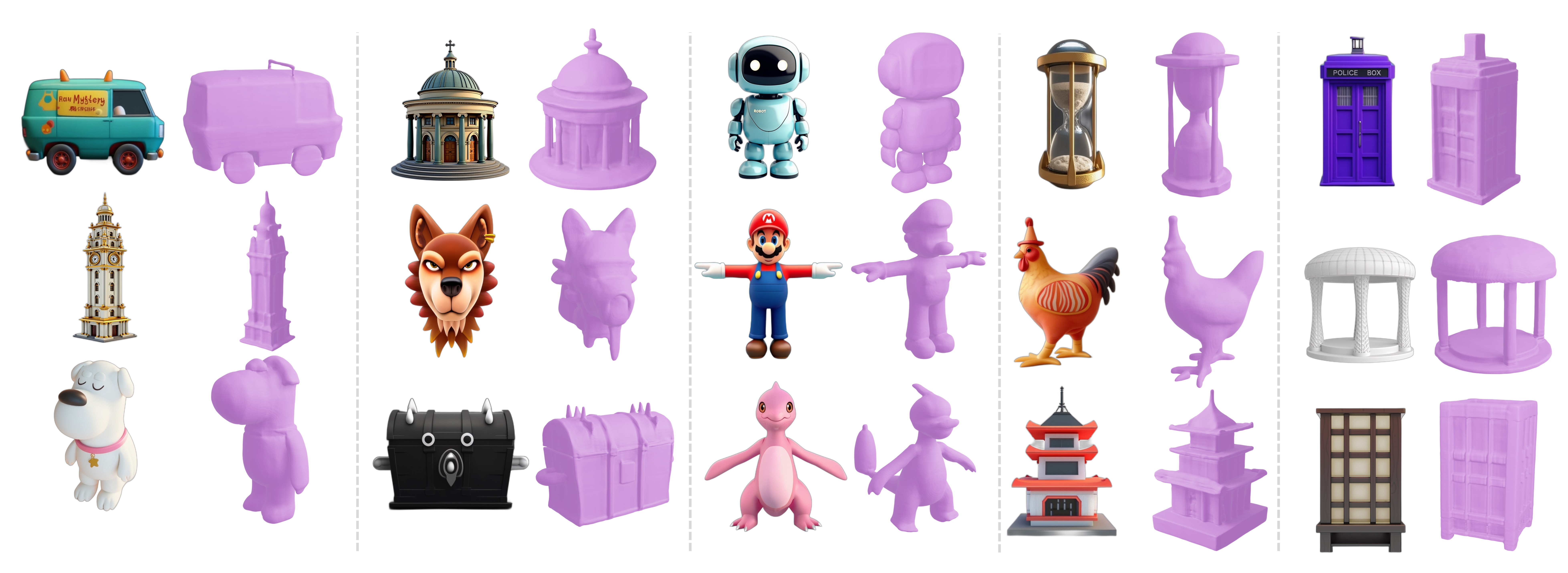} 
\caption{
VAT enables a robust and generalizable 3D generation conditioned on in-the-wild images.
}
\label{fig: our_visualization}
\end{figure*}

\subsection{Implementation details}


The input point cloud in VAT consists of 80,000 points uniformly sampled from the Objaverse dataset~\cite{deitke2023objaverse}. These points are transformed into 1D features, resulting in a length $L = 3072$ and channel dimension $C = 768$. The encoder for in-context compression includes 12 self-attention layers. The length $K$ of the compressed tokens varies from 256 to 1024, depending on the compression ratio. 
%
Initially, we train VAT for 200,000 steps without quantization, followed by fine-tuning all parameters, including codebook parameters, for an additional 100,000 steps. 
The decoder in VAT de-tokenization phase comprises one cross-attention layer and 12 self-attention layers with the same channel dimension as the encoder. 
For supervision, we sample 20,000 uniform points and 20,000 near-surface points during training. 
%
The next-scale AR model follows the architecture of VAR~\cite{VAR}.
We select 200,000 high quality data in Objaverse~\cite{deitke2023objaverse} for training.
The model utilizing 1,024 compressed tokens contains 0.5 billion parameters and was trained for one week on 96 NVIDIA H20 GPUs with 96GB of memory.
Additional training and architecture details can be found in the supplements.

\section{Experiment}

\subsection{Experiment Setup}
To evaluate the reconstruction accuracy of the first stage of the tokenizer, we use Occupancy Accuracy (Acc.) and Intersection-over-Union (IoU) as our primary metrics, which are computed based on occupancy predictions from 40,000 randomly sampled query points in 3D space, along with an additional 40,000 points sampled near the surface.
We randomly select 500 3D meshes from the Objaverse dataset~\cite{deitke2023objaverse} as our evaluation dataset, covering a wide variety of object shapes. Each shape is normalized to fit within its bounding box. The absolute occupancy value is then calculated based on the distance to the closest triangle of the surface. The sign of the occupancy value is determined by checking whether the point is inside or outside the surface, following the operation in NGLOD~\cite{takikawa2021nglod}.
To further assess the model's ability to capture fine details, we introduce Near-Surface Accuracy (Near-Acc.), which is the prediction accuracy of 10,000 points located within a distance of 0.05 from the GT surface.


To obtain the mesh, we sample query points on a grid with a resolution of $256^3$ and reconstruct the shapes using the Marching Cube~\cite{10.1145/37402.37422, shen2023flexible}.
Subsequently, Chamfer Distance (Cham.) and F-score (with a threshold of 0.01) are used to evaluate mesh quality in the second stage of generation based on the image condition. These metrics are calculated between two point clouds, each containing 10,000 points, sampled from the reconstructed and ground-truth surfaces. Since the generated mesh may not be perfectly aligned with the ground-truth mesh, we apply the Iterative Closest Point (ICP) algorithm to align the reconstructed surface with the ground-truth surface by minimizing the point-to-point distance between corresponding points.
%

\subsection{State-of-the-art 3D Generation}
\begin{table*}[h]
\centering
\setlength{\tabcolsep}{3pt}
\begin{tabular}{ccc|ccc|ccc}
\toprule
\multirow{2}{*}{Type} & \multirow{2}{*}{Method} & \multirow{2}{*}{Name} & 
 \multicolumn{3}{c|}{Objaverse} & \multicolumn{3}{c}{GSO} \\
 
& & & IoU(\%)$\uparrow$ & Cham.$\downarrow$ & F-score(\%)$\uparrow$ & IoU(\%)$\uparrow$ & Cham.$\downarrow$ & F-score(\%)$\uparrow$ \\
\midrule
\multirow{4}{*}{LRM} & \multirow{3}{*}{NeRF} & Triposr & 72.6 & 0.023 & 58.2 & 75.8& \textbf{0.011} & 62.0\\
 & & InstantMesh~\cite{xu2024instantmesh} & 68.7 & 0.029 & 58.3 & 61.9& 0.021& 51.6\\
 & & CRM~\cite{wang2024crm} & 76.3 & 0.020 & 61.4 & 72.4& 0.010& 60.8\\
 & Gaussian & LGM~\cite{tang2024lgm} & 67.6 & 0.025 & 49.3 & 71.0& 0.013& 53.2\\
\cline{1-3}
\multirow{7}{*}{\begin{tabular}[c]{@{}c@{}}3D\\Generation\end{tabular}} 
& \multirow{4}{*}{Diffusion} & Michelangelo~\cite{zhao2024michelangelo} & 74.5 & 0.028 & 62.5 & 65.3& 0.018& 52.3\\
 & & Shap-e~\cite{jun2023shap} & 66.8 & 0.029 & 46.3 & 64.8& 0.019& 49.1\\
 & & CraftsMan~\cite{li2024craftsman} & 72.2 & 0.021 & 56.1 & 69.4& 0.011& 55.2\\
 & & CLAY (0.5B)* ~\cite{zhang2024clay} & 77.1& 0.021& 63.4 & 71.7& 0.010& 60.8\\
 
\cline{2-3}
& \multirow{3}{*}{\begin{tabular}[c]{@{}c@{}}AR\\Modeling\end{tabular}}

 & G3PT (0.5B)~\cite{g3pt} & 82.11 & 0.015 & 75.1 & 74.5 & 0.014& 64.2\\
\cline{3-9}
& & {VAT-M (0.5B) } & {88.7}& {0.013}& {83.9} &83.1 & 0.012 & \textbf{70.1}\\
& & \textbf{VAT-L (0.5B)} & \textbf{90.1}& \textbf{0.013}& \textbf{86.6} & \textbf{84.2} & 0.013 & 68.2\\
\bottomrule
\end{tabular}
\caption{Comparison of state-of-the-art 3D generation methods. (*: Reproduction)}
\label{tab:comparison}
\end{table*}

The quantitative comparisons are presented in Table~\ref{tab:comparison} on two dataset, Objaverse~\cite{deitke2023objaverse} and GSO~\cite{radford2021learning}. The evaluated methods include LRM-based approaches such as InstantMesh~\cite{xu2024instantmesh} and CRM~\cite{wang2024crm}
Triposr~\cite{tochilkin2024triposr} 
maps image tokens to implicit 3D triplanes under multi-view image supervision, while LGM~\cite{tang2024lgm} replaces the triplane NeRF representation with 3D Gaussians~\cite{kerbl20233d} to improve rendering efficiency.
Additionally, diffusion-based methods such as Michelangelo~\cite{zhao2024michelangelo}, Shap-E~\cite{jun2023shap}, CraftsMan~\cite{li2024craftsman}, and CLAY~\cite{zhang2024clay} are compared. For AR modeling, we follow the architecture of G3PT~\cite{g3pt}, which is a scalable next-scale autoregressive framework.
The results highlight our significant advantage, which outperforms all other methods with a substantial margin in all metrics, demonstrating superior generation quality and fidelity.



\begin{figure}[t]
\centering
\includegraphics[width=0.45\textwidth]{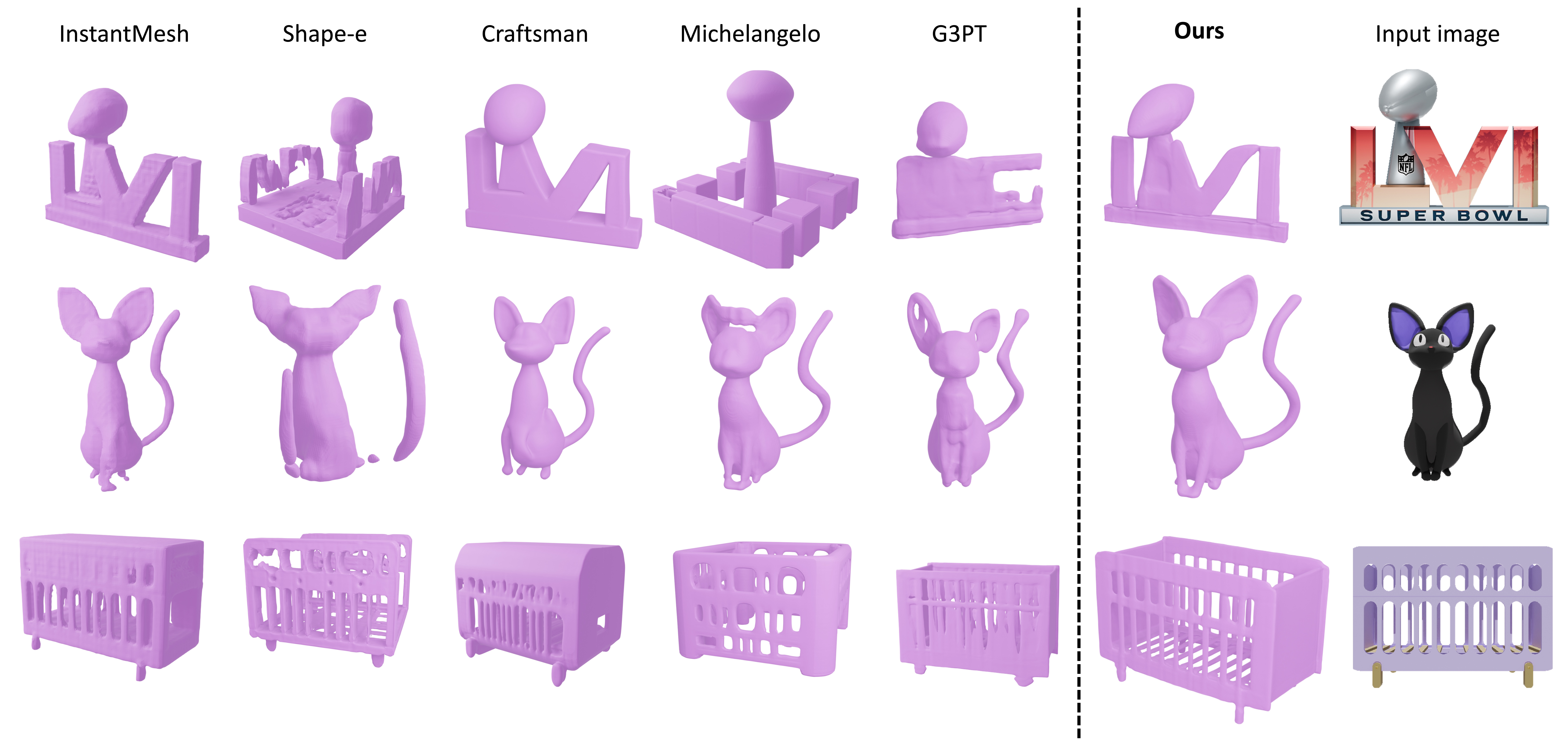} 
\caption{
Qualitative comparision of state-of-the art 3D generation methods in Objaverse dataset. }
\label{fig: objav_compare}
\end{figure}

As shown in Fig.~\ref{fig: wild_compare} and Fig.~\ref{fig: objav_compare}, we perform qualitative comparisons with other state-of-the-art methods on images from the Objaverse dataset and in-the-wild images for the image-to-3D task. 
%
LRM-based methods generate 3D models that closely resemble the input images but often exhibit noise and mesh artifacts. Diffusion-based methods, such as Michelangelo, produce plausible geometry but struggle to maintain alignment with the semantic content of the conditional images. 
Our method achieves a superior balance between quality and realism. Furthermore, our VAT enables generation of smoother and more intricate geometric details compared to G3PT~\cite{g3pt}.

\subsection{Main Properties}

\textbf{Curse of Hierarchy.}
This experiment demonstrates that naively increasing token numbers does not inherently enhance reconstruction performance, as shown in Table~\ref{tab:size_token}. Instead, excessive tokenization can degrade cross-scale consistency and reconstruction fidelity, a phenomenon we term the ``curse of hierarchy". This experiments are conducted on various latent token number without employing in-context compression and VVQ, which shares the same structure illustrated in Figure~\ref{fig: method_vq}(a).
To evaluate the reconstruction performance of each tokenizer, we use 
Cross-scale IoU (CS-IoU) to assess semantic consistency across token scales, which are measured at each scale $s$ by dropping tokens beyond scale $s$ and averaging performance across all scales.
Table~\ref{tab:size_token} shows that model performance with naive implementation peaks at 1024 tokens, achieving an optimal balance between accuracy and cross-scale consistency. Beyond this point, adding more tokens leads to fragmentation, which disrupts the hierarchical structure and reduces overall performance.
In contrast, in-context compression significantly improves reconstruction results, even with far fewer tokens. However, semantic consistency drops substantially without VVQ. By incorporating VVQ, our VAT achieves the best balance between reconstruction accuracy and cross-scale consistency.

\begin{table}[t]
    \centering
        \setlength{\tabcolsep}{2pt}
        \begin{tabular}{cc|cc|ccccc}
            \toprule
            Comp. & VVQ & \#Token & \#Scale & Acc.(\%) & IOU(\%) & CS-IOU.(\%) \\
            \midrule
            $\times$& $\times$& 256 & 10 & 82.14& 55.73 & 32.45\\
            $\times$& $\times$& 576 & 11 & 86.45& 63.13&  {40.57}\\
            $\times$& $\times$& 1024 & 12 & 88.12 & 65.86&  33.15\\
            $\times$& $\times$& 2408 & 13 & {89.32} & {68.57} & 29.31 \\
            $\times$& $\times$& 3072 & 14 &  80.14&  50.18&    28.40\\
            \midrule
            $\checkmark$& $\times$& 576 & 11 & 91.45& \textbf{73.12} & 15.12\\
            $\checkmark$& $\checkmark$& 576 & 11 & \textbf{91.73} & 72.34 & \textbf{47.32}\\
            \bottomrule
        \end{tabular}
        \caption{
        Reconstruction results with varying numbers of tokens, with and without in-context token compression (Comp.) and Variational Vector Quantization (VVQ). 
       }\label{tab:size_token}
\end{table}


\begin{table*}[t]
    \centering
        \setlength{\tabcolsep}{3pt}
        \begin{tabular}{c|cccc|ccc}
            \toprule
            \multirow{3}{*}{\begin{tabular}[c]{@{}c@{}}Training\\ Strategy\end{tabular}} & \multicolumn{4}{c|}{Reconstruction} & \multicolumn{3}{c}{Generation}   \\
            & Acc.(\%) $\uparrow$        & IOU(\%) $\uparrow$        & CS-Acc.(\%)  $\uparrow$        & CS-IOU(\%)  $\uparrow$     & {\begin{tabular}[c]{@{}c@{}}Training\\ loss\end{tabular}}    & {\begin{tabular}[c]{@{}c@{}}F-score(\%) $\uparrow$\\  (all scale)\end{tabular}}  & {\begin{tabular}[c]{@{}c@{}}F-score(\%) $\uparrow$\\  (last scale)\end{tabular}}      \\

            \midrule
            None & 91.45 & \textbf{73.12} & 77.02& 15.12 &  \textbf{0.98}& 64.15& 91.13\\
            Dropout~\cite{li2024imagefolderautoregressiveimagegeneration} & 89.23 & 64.15 & 82.34 & 36.31 & 1.02& 67.15& 89.94\\
            Stochastic~\cite{lee2022autoregressive} & 88.12 & 61.75& 80.93 & 32.14 & 1.13& 70.42& 90.02\\
            \textbf{VVQ (Ours)} & \textbf{91.73} & 72.34 & \textbf{85.58} & \textbf{47.32} & 1.08 & \textbf{83.92} & \textbf{91.23} \\
            
            \bottomrule
        \end{tabular}
       \caption{Comparison of reconstruction and the generation performance using  tokenizers trained by different strategy.
       Here, ``None" refers to VAT without adding Gaussian noise in VVQ. 
       }\label{tab:sample_compare}
       \vspace{-5pt}
\end{table*}

\begin{figure}[ht]
\centering
\includegraphics[width=0.5\textwidth]{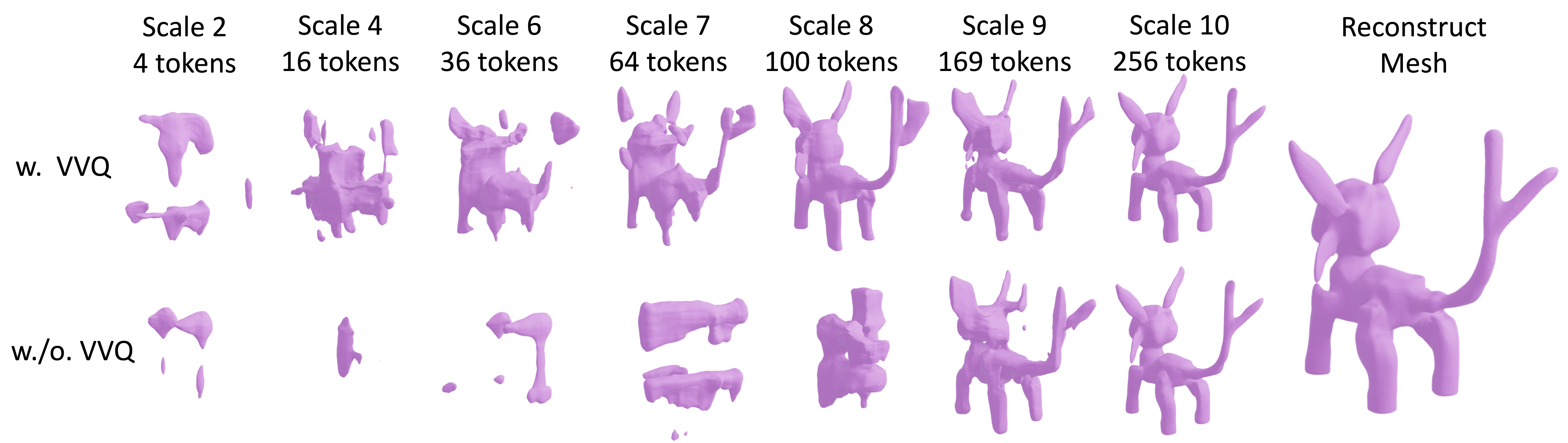} 
\caption{
Visualization of reconstructed mesh from different scales of tokens.
}
\label{fig: multiscale_vis}
\vspace{-5pt}
\end{figure}
\textbf{Necessity of VVQ.}
As shown in Table~\ref{tab:sample_compare}, we compare VVQ with three alternative tokenization methods designed to enhance interconnections among token maps: (1) Dropout~\cite{li2024imagefolderautoregressiveimagegeneration}, which randomly drops the last few scales of tokens during the tokenizer's training, (2) Stochastic Sampling~\cite{lee2022autoregressive}, which applies probabilistic sampling of the code map to reduce discrepancies between training and inference, and (3) None, which applies no interconnection technique. All methods were trained and evaluated under the same network architecture and training parameters for a fair comparison.
For generation performance, we separately train four separate AR models, each conditioned on a different tokenizer, and measure the final F-score of the generated mesh based on the same image input conditions. Additionally, we assess generation performance at the last two scales by providing ground-truth token maps for the first 10 layers, generating only the last two layers of tokens.

As shown in Table~\ref{tab:sample_compare}, all methods show similar Accuracy and IoU, but Cross-scale metrics (CS-Acc. and CS-IoU) highlight VVQ's advantage, indicating that VVQ effectively captures hierarchical inter-scale relationships.
While all the AR model are all well-trained with similar training loss, final generation quality shown in F-score of all scales varies significantly. With ground-truth tokens for the first 10 scales, generation quality becomes more consistent, highlighting that other methods without VVQ suffer from exposure bias, where training-inference discrepancies cause cumulative errors in AR modeling. VVQ mitigates this by projecting token maps into a shared Gaussian distribution, smoothing the token distribution and enhancing consistency across scales. Fig.~\ref{fig: multiscale_vis} visualizes reconstructed meshes at different scales with and without VVQ.


\begin{table}[htb]
    \centering
    \scriptsize
    \setlength{\tabcolsep}{2pt}
    \begin{tabular}{cc|ccccc}
        \toprule
        Method & \#Token &  Acc.(\%) & IOU(\%)  & Near-Acc.(\%) \\
        \midrule
        \multirow{2}{*}{Cross-attention} & 576 & 87.93 & 64.25 & 58.56  \\
                             & 1024 & 90.51 & 68.42 & 60.90 \\
        \midrule
        \multirow{2}{*}{\textbf{Triplane (Ours)}} & 576 & 91.73 & 72.34 & 64.58  \\
                               & 1024 & \textbf{92.31} & \textbf{74.71} & \textbf{67.20}  \\
        \bottomrule
    \end{tabular}
    \caption{Performance comparison of different decoding structures.}
    \vspace{-2em}
    \label{tab:cqt_comparison}
\end{table}


\begin{figure}[h]
\centering
\includegraphics[width=0.35\textwidth]{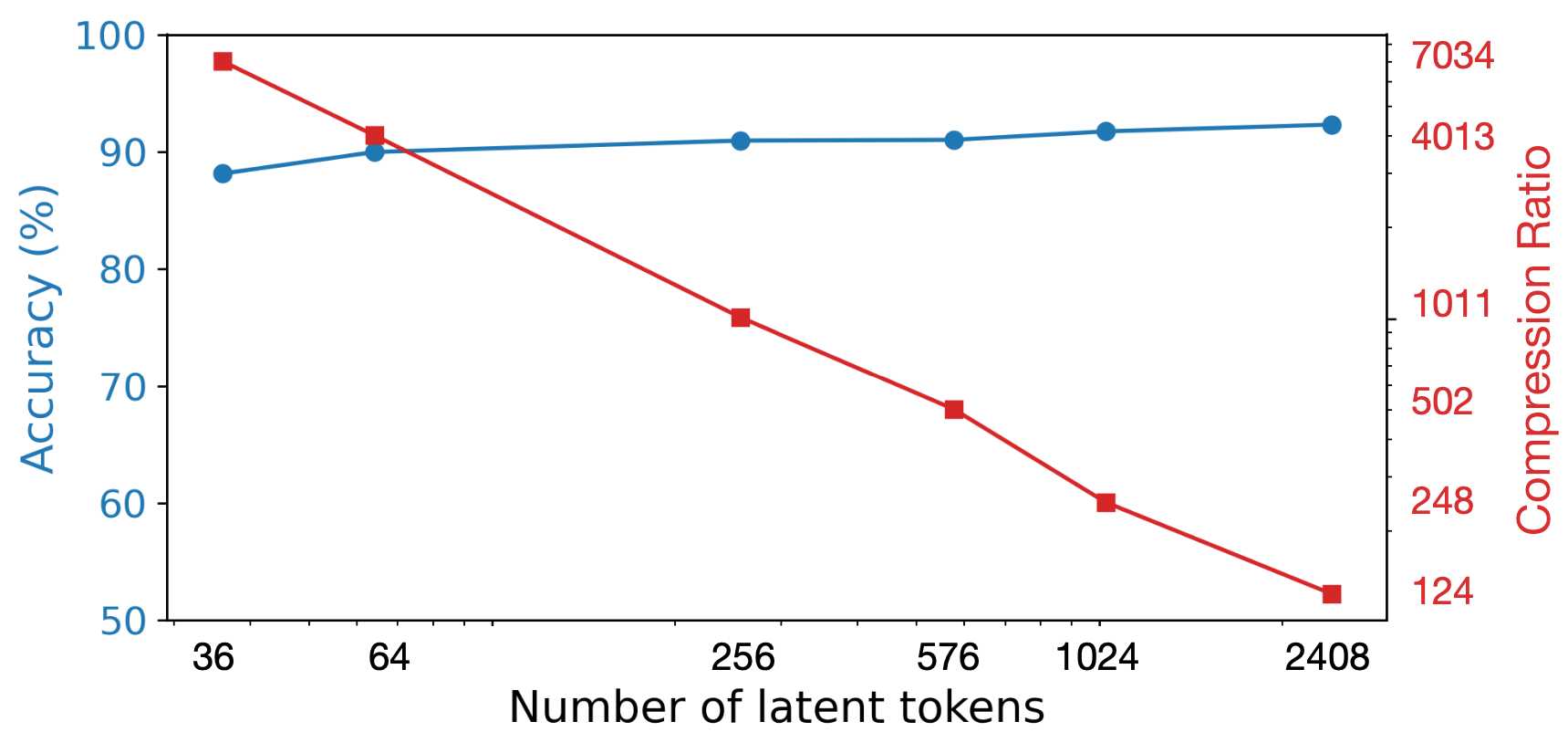} 
\caption{
Compression ratio with different VAT variants. }
\label{fig: compress_plot}
\end{figure}


\textbf{Compression.} We compare several VAT variants with different latent token sizes $K$, ranging from 36 to 2408. The compression ratio is calculated as the size of the original mesh (after simplification) divided by the storage size of our token representation. Since each token can be represented by a 2-bit integer, the size of our latent representation is computed by multiplying the total number of tokens across all scales by 2.
As shown in Fig.~\ref{fig: compress_plot}, although reconstruction accuracy progressively improves as the number of latent tokens increases, significant enhancements are predominantly observed once $K$ exceeds 200. When the latent token count reaches 256, VAT achieves a substantial compression ratio of approximately 4000.


\subsection{Ablation study}

\textbf{Compression strategy.} 
As shown in Table~\ref{tab:compress_strategy}
we ablate different token compression strategies used in VAT.
The ``Pooling'' approach discards latent tokens and applies one-dimensional pooling directly to the feature outputs, as shown in Fig.~\ref{fig: method_vq}(a). With an input feature token size of 3072 and pooled token size of 1024, this method simplifies the architecture but limits the model's ability to capture complex spatial details, leading to reduced performance.
Next, we evaluate ``Q-Former''~\cite{li2023blip}, which uses one layer of cross-attention between latent tokens and 3D input features for token compression, which still underperforms compared to our In-context Transformer. 

\begin{table}[t]
    \centering
    \scriptsize
        \setlength{\tabcolsep}{5pt}
        \begin{tabular}{c|ccc}
            \toprule
            {\begin{tabular}[c|]{@{}c@{}}Compression\\ Strategy\end{tabular}} & Acc.(\%) & IOU(\%)  & Near-Acc.(\%) \\

            \midrule
            {Pooling} & 87.89 & 64.42 & 61.25\\
            Q-Former & 90.43 & 67.12 & 62.78\\
            \textbf{In-context (Ours)} & \textbf{91.73} & \textbf{72.23} & \textbf{64.58}\\
            
            \bottomrule
        \end{tabular}
       \caption{Ablation study on different compression strategy.}\label{tab:compress_strategy}
       \vspace{-5pt}
\end{table}





\textbf{Triplane architecture}.
%
As shown in Table~\ref{tab:cqt_comparison}, the Triplane architecture demonstrates superior performance metrics across all evaluation criteria compared with a Cross-attention mechanism~\cite{DBLP:journals/tog/ZhangTNW23}, which replaces the Triplane with a single Cross-attention layer. 
These findings underscore the superiority of the Triplane architecture in delivering high-fidelity reconstruction.

\section{Conclusion}
In this paper, we introduce the Variational Tokenizer (VAT) as an innovative solution to the challenges of compact 3D representation and autoregressive 3D generation. Unlike traditional tokenizers, which are designed for 2D images and leverage inherent spatial sequences and multi-scale relationships, 3D data lacks a natural order, complicating the task of compressing it into manageable tokens while preserving its structural details. VAT addresses this challenge by transforming unordered 3D data into subspaces of a Gaussian distribution, enabling efficient and effective autoregressive generation.

{
    \small
    \bibliographystyle{vat}
    \bibliography{vat}
}

\clearpage
\setcounter{page}{1}
\maketitlesupplementary

\section{More implementation details}

\subsection{Dataset preparation}


Our training dataset is derived from the Objaverse dataset, which contains around 800k 3D models created by artists~\cite{deitke2023objaverse}. To ensure high-quality training data, we applied a rigorous filtering process. Specifically, we removed objects that: (i) lack texture maps, (ii) occupy less than 10\% of any rendered view, (iii) consist of multiple separate objects, or (iv) exhibit low-quality geometry, such as thin structures, holes, or texture-less surfaces. This filtering reduced the dataset to approximately 270k high-quality instances.

For each selected object, we normalized it to fit within a unit cube. In addressing the occupancy field extraction for non-watertight meshes, we employed a standardized geometry remeshing protocol. Specifically, we utilized the Unsigned Distance Field (UDF) representation for the mesh, inspired by CLAY~\cite{DBLP:journals/tog/ZhangWZQPJYXY24}, and determined whether the grid points are "inside" or "outside" based on observations from multiple angles.

\begin{figure}[ht]
\centering
\includegraphics[width=0.4\textwidth]{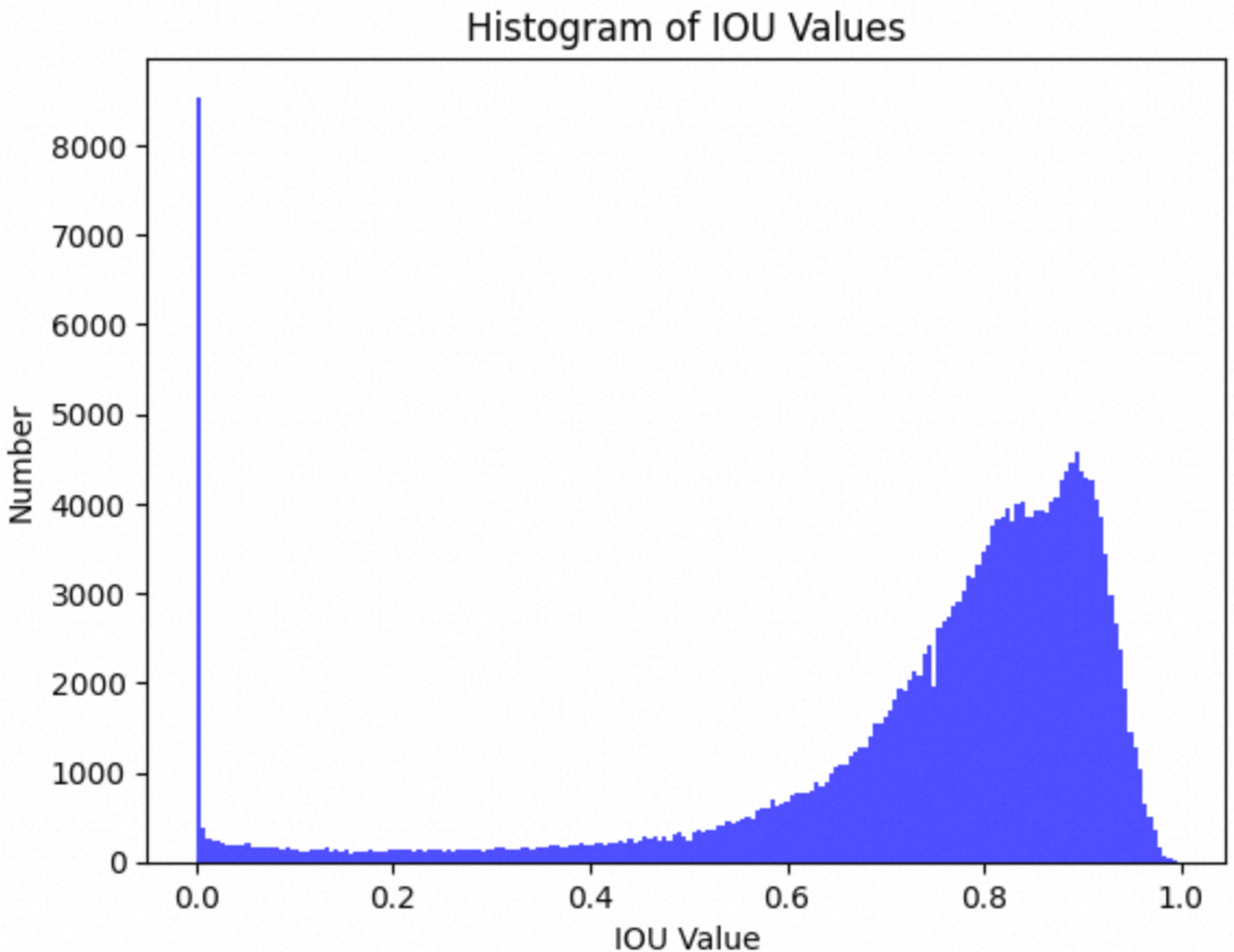} 
\caption{IoU distribution histogram of a tiny VAT (256 tokens) on the Objaverse dataset. Data with IoU greater than 0.2 is selected for the second stage of training.}
\label{fig: plot_iou}
\end{figure}

To further refine the dataset, we used a pre-trained tiny VAT model (256 latent tokens) to predict IoU for each instance, as shown in Fig.~\ref{fig: plot_iou}. Objects with an IoU of 0 were discarded. For training larger VAT models (512/1024 tokens), we only used instances with IoU above 0.2. In the second stage of AR modeling, we further refined the dataset by selecting only those with IoU greater than 0.4.

Similar to SV3D~\cite{DBLP:conf/eccv/VoletiYBLPTLRJ24}, we generate a 24-frame RGBA orbit at a resolution of 512×512 using Blender’s EEVEE renderer. Our camera is set with a field-of-view of 33.8 degrees. For each object, we dynamically position the camera at a distance that ensures the rendered object fills the image frame effectively and consistently, without being cut off in any perspective. The camera starts at an azimuth of 0 degrees for each orbit and is placed at a randomly selected elevation within the range of -5 to 30 degrees. The azimuth angle increases by a fixed increment of $\frac{360}{24}$ degrees between each frame. We randomly selected one rendered image and utilize a white background color for training.

We emphasize the precise textual prompts within our 3D model to effectively capture the geometric and stylistic details of objects. To this end, we crafted distinctive prompt tags~(e.g. "symmetric geometry", "asymmetric geometry", “sharp geometry”, "smooth geometry", "low-poly geometry", "high-poly geometry", "simple geometry", "complex geometry", "single object", "multiple object") and employed GPT-4V to generate detailed annotations.  
This method significantly enhances the model's ability to interpret and generate complex 3D geometric shapes with subtle details and a broad range of styles.

\subsection{VAT architecture}

\begin{figure*}[t]
\centering
\includegraphics[width=0.9\textwidth]{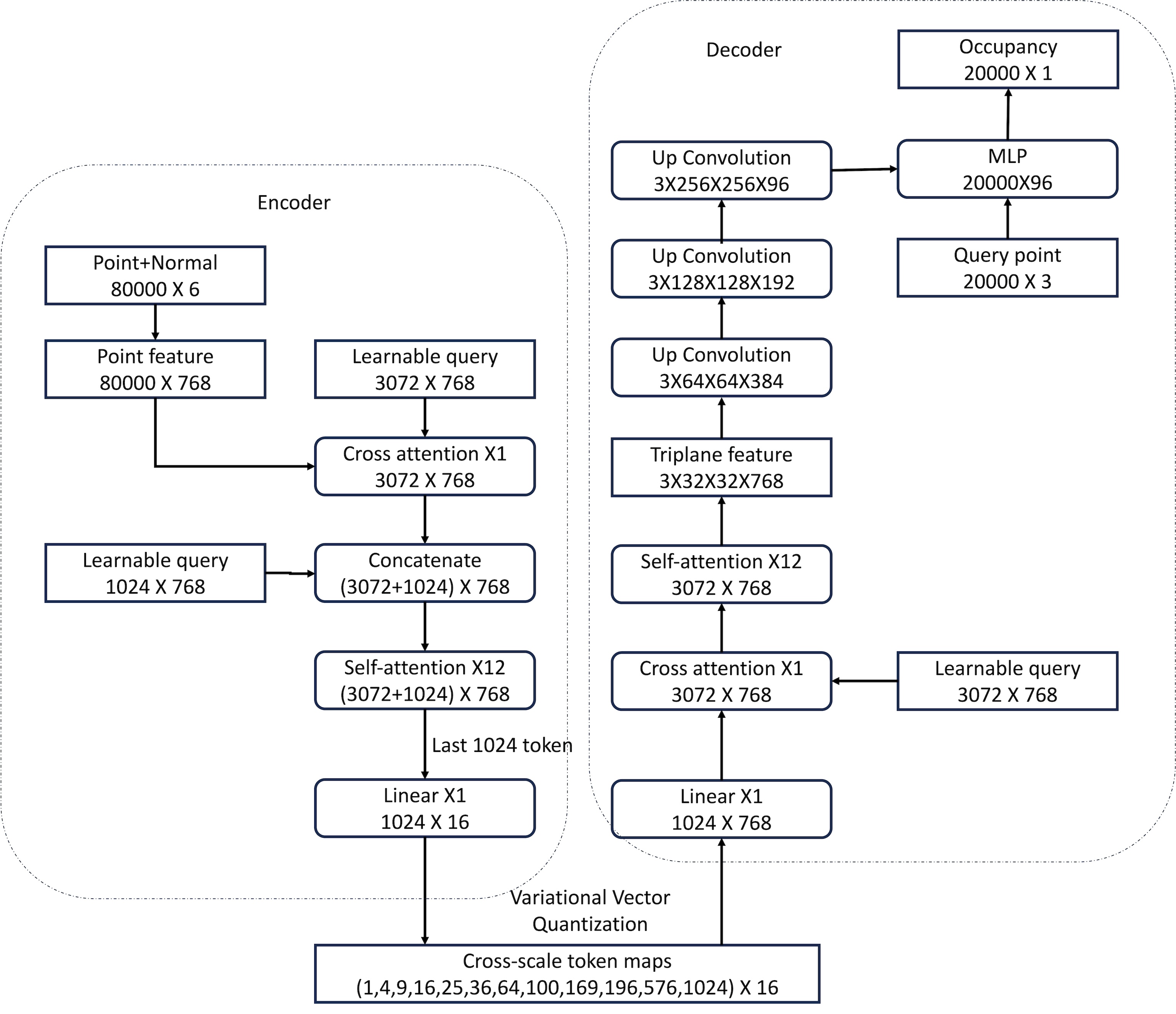} 
\caption{Detailed network architecture of VAT.}
\label{fig: fig_supp_network_arch}
\end{figure*}

The input point cloud in VAT consists of 80,000 points uniformly sampled from the Objaverse dataset~\cite{deitke2023objaverse},  which include normalized positions and normals for each point. As shown in Fig.~\ref{fig: fig_supp_network_arch},  we enhance the spatial encoding of these points using Fourier features~\cite{jaegle2021perceivergeneralperceptioniterative}, capturing intricate geometric structures.
These points are transformed into 1D features using a cross-attention layer with $L = 3072$ learnable queries, resulting in a length $L = 3072$ and channel dimension $C = 768$. Specifically, a set of learnable tokens ${I_p} \in \mathbb{R}^{3072 \times 768}$ queries these point cloud features through cross-attention, embedding 3D information into latent features.
Then, 1024 tokens are concatenated with the 3072 features as the input of 12 self-attention layers. The output of the encoder only keep the 1024 tokens for compression. 
Before the VVQ, a linear layer projects and unprojects the features into a lower-dimensional space of $C_q = 16$. Initially, we train VAT for 200,000 steps without quantization, followed by fine-tuning all parameters, including codebook parameters, for an additional 100,000 steps. The vocabulary size of the codebook is set to 2048 and 16,384 depending on the accuracy requirement. The decoder in VAT de-tokenization phase comprises one cross-attention layer and 12 self-attention layers with the same channel dimension as the encoder. 

An explicit triplane latent representation is employed to convert the latent feature ${\hat{I}}$ into 3D geometry~\cite{wu2024direct3d, wang2022rodingenerativemodelsculpting}. This process reshapes ${\hat{I}}$ into three 2D planes, yielding ${I_{tri}} \in \mathbb{R}^{3 \times r \times r \times D}$. Convolutional layers then progressively upsample ${I_{tri}}$, generating high-resolution triplane features, denoted as $\mathbf{T} = (\mathbf{T}_{XY}, \mathbf{T}_{YZ}, \mathbf{T}_{XZ})$. This approach efficiently captures intricate 3D spatial details.

However, directly upsampling the triplane often leads to blurring and aliasing artifacts at high resolutions due to neglecting the sampling area~\cite{barron2021mipnerf}. To address this, each triplane is represented using three mipmaps, each with progressively higher resolutions upsampled from ${I_{tri}}$ via convolutional layers that double in size (i.e., with three different resolutions: $r, r/2, r/4$). Subsequently, an MLP-based mapping network interpolates features from these three triplanes $\mathbf{T}$ at different levels, concatenating all features to predict occupancy values.

\subsection{Training details}

\subsubsection{Supervision signal in Stage 1}
A semi-continuous approach is adopted to reduce abrupt gradient changes near the object surface, enhancing the stability of model training. For a query point $\mathbf{x}$, occupancy values are binary for points beyond $s = \frac{1}{128}$ from the surface, while continuous values are assigned to points within this range, facilitating smoother gradient flow:
\[
o(\mathbf{x}) = 
\begin{cases} 
1, & \text{if } \text{sdf}(\mathbf{x}) < -s \\
0.5 - \frac{0.5 \cdot \text{sdf}(\mathbf{x})}{s}, & \text{if } -s \leq \text{sdf}(\mathbf{x}) \leq s \\
0, & \text{if } \text{sdf}(\mathbf{x}) > s 
\end{cases}
\]
where $\text{sdf}(\mathbf{x})$ is the Signed Distance Function (SDF) of $\mathbf{x}$, helping maintain training stability around the surface boundary.

For supervision, we sample 20,000 uniform points and 20,000 near-surface points during training. The AdamW optimizer is employed with a learning rate of $1 \times 10^{-4}$, and the model is trained on 8 NVIDIA A100 GPUs with a batch size of 256. 

\subsubsection{Model setup and hyperparameters in Stage 1}

    \begin{itemize}
        \item \textbf{VAT input:} Point cloud, $80000$ points.
        \item \textbf{Base channels:} 768.
        \item \textbf{Number of self-attention blocks:} 12.
        \item \textbf{Latent tokens:} $64/256/1024$.
        \item \textbf{Vocabulary size:} $2048/16384$.
        \item \textbf{Occupancy loss weight:} 1.0.
        \item \textbf{Codebook MSE weight:} 0.2.
        \item \textbf{KL regularization loss weight:} $10^{-4}$.
        \item \textbf{Peak learning rate:} $10^{-4}$.
        \item \textbf{Learning rate schedule:} Linear warm-up and cosine decay.
        \item \textbf{Optimizer:} Adam with $\beta_1 = 0.9$ and $\beta_2 = 0.99$.
        \item \textbf{EMA model decay rate:} 0.99.
        \item \textbf{Batch size:} 256.
    \end{itemize}

\subsubsection{Model setup and hyperparameters in Stage 2}

\begin{figure*}[ht]
\centering
\includegraphics[width=0.9\textwidth]{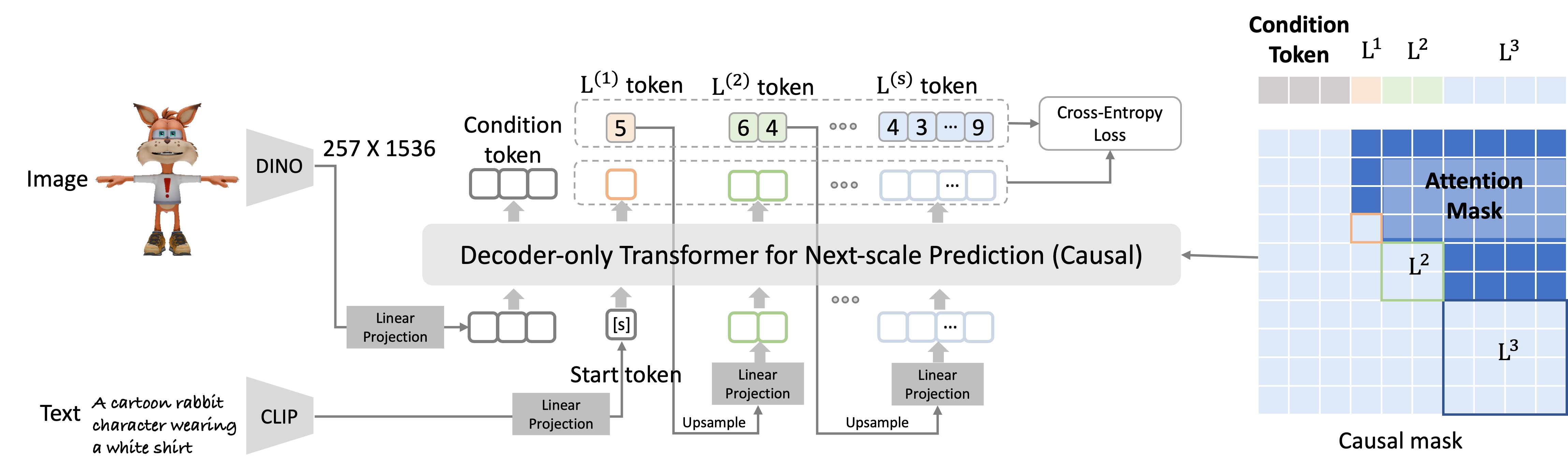} 
\caption{Network architecture for training AR model in stage 2.}
\label{fig: fig_supp_ar_arch}
\end{figure*}

As shown in Fig.~\ref{fig: fig_supp_ar_arch}, we adopt the architecture of standard decoder-only transformers akin to GPT-2 with adaptive normalization (AdaLN). For text-conditional synthesis, we use the text embedding as the start token [s] and also the condition of AdaLN. We
use normalized queries and keys to unit vectors before attention. We adapt learnable queries as the position embedding. 

    \begin{itemize}
        \item \textbf{Token number of each scale:}
        
        (1,4,9,16,25,36,64,100,169,196,576,1024).
        \item \textbf{Base channels:} 1280.
        \item \textbf{Number of self-attention blocks:} 12.
        \item \textbf{Peak learning rate:} $10^{-4}$.
        \item \textbf{Learning rate schedule:} Linear warm-up and cosine decay.
        \item \textbf{Optimizer:} Adam with $\beta_1 = 0.9$ and $\beta_2 = 0.99$.
        \item \textbf{Batch size:} 1600.
    \end{itemize}

\section{More Visualizations}
\subsection{Distribution of the codebook in VVQ}



\begin{figure}[h]
\centering
\includegraphics[width=0.4\textwidth]{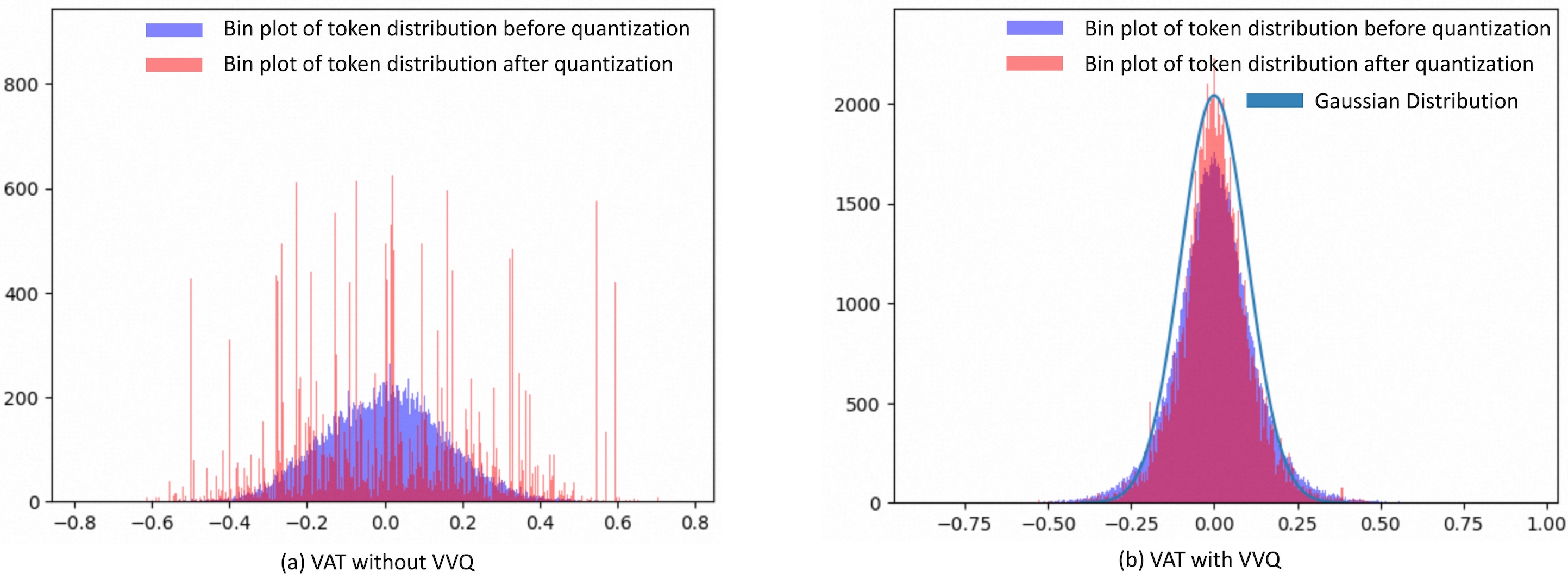} 
\caption{Comparison of token distribution before and after quantization using (a) VAT without VVQ and (b) VAT with VVQ. The blue histogram represents the token distribution before quantization, while the red histogram shows the distribution after quantization. Additionally, in Figure 1(b), the Gaussian distribution is overlaid for comparison.}
\label{fig: fig_supp_bin_plot}
\end{figure}

In Fig.~\ref{fig: fig_supp_bin_plot}, we visualize the distribution of token features before and after quantization given two VAT variants. Specifically, in Fig.~\ref{fig: fig_supp_bin_plot}(a), we employ the tokenizer without VVQ. 
%
For the distribution shown in Fig.~\ref{fig: fig_supp_bin_plot}(b), we present the pre-quantization feature distribution of $Z_0$ (adding Gaussian noise) in blue and the dequantized output $\hat{Z}$ in red.
%
This plot clearly demonstrates that when VVQ is utilized, the distribution of discrete tokens conforms to a Gaussian distribution. In contrast, without the introduction of VVQ, the distribution of discrete tokens exhibits significant deviation from the pre-quantization state, leading to a more complex distribution. 

\begin{figure*}[t]
\centering
\includegraphics[width=0.9\textwidth]{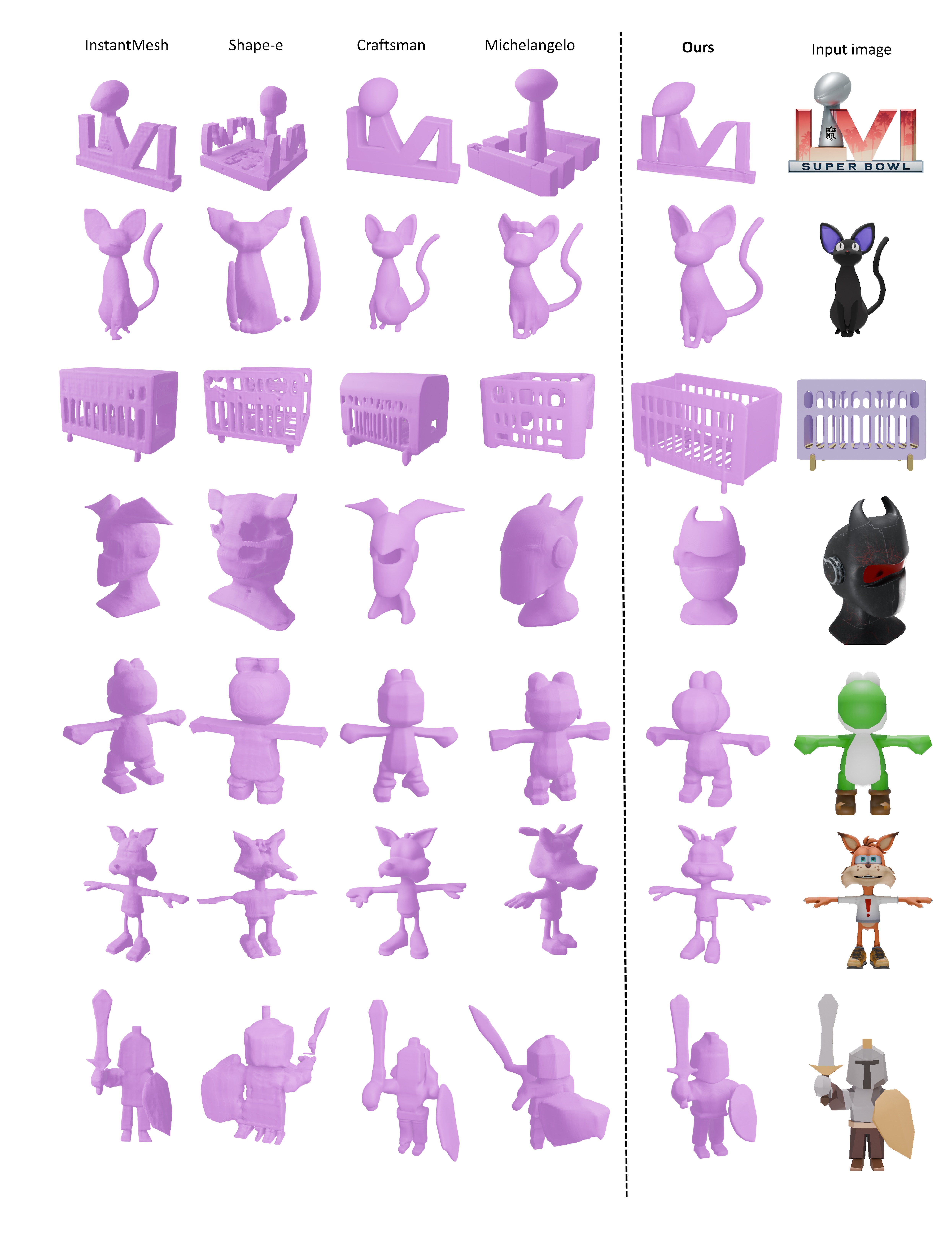} 
\caption{Qualitative comparision of state-of-the art 3D generation methods in Objaverse dataset.}
\label{fig: fig_supp_compare_1}
\end{figure*}


\begin{figure*}[t]
\centering
\includegraphics[width=0.9\textwidth]{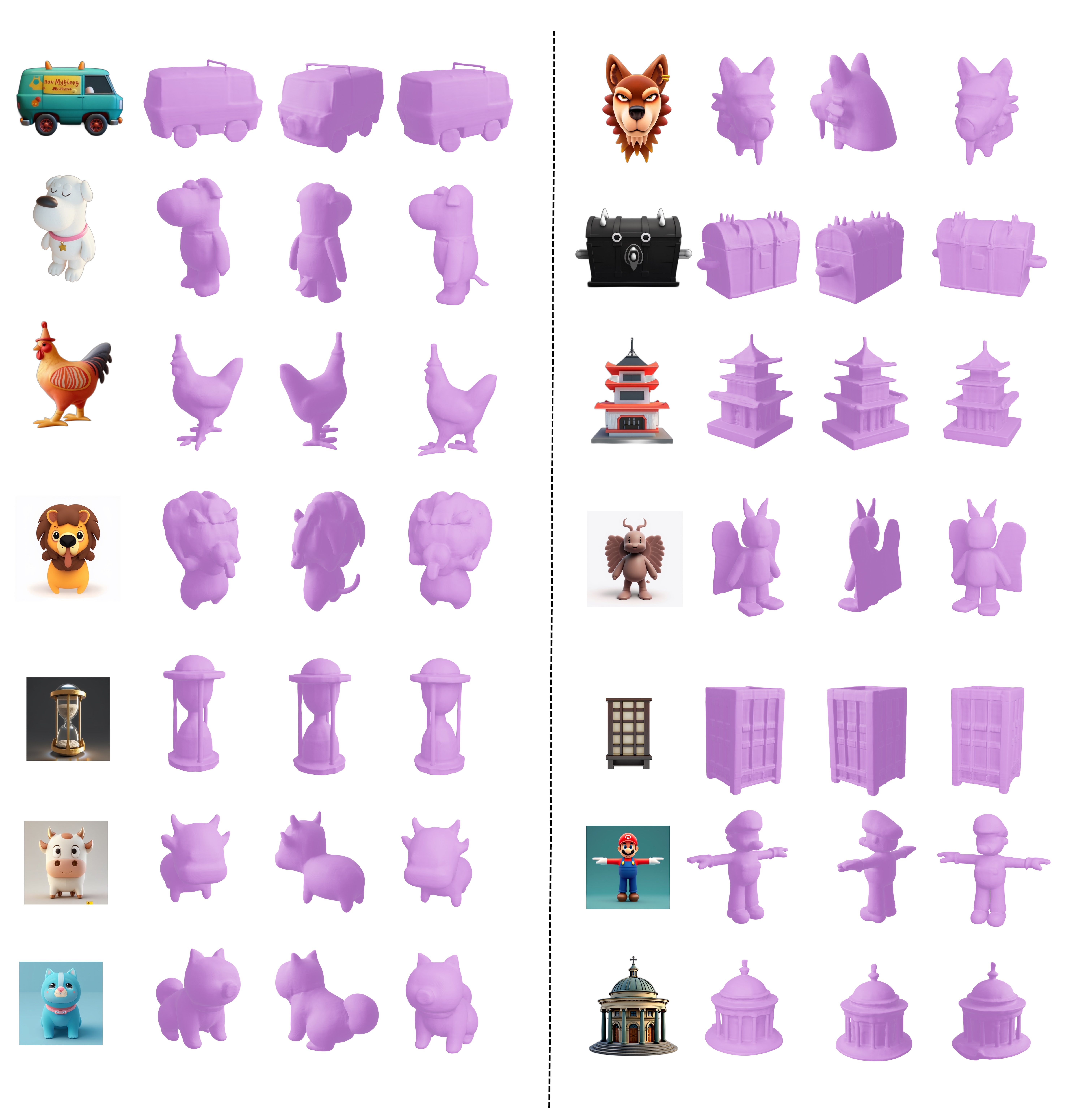} 
\caption{More Visualizations.}
\label{fig: fig_supp_compare_wild}
\end{figure*}

\begin{figure*}[t]
\centering
\includegraphics[width=0.8\textwidth]{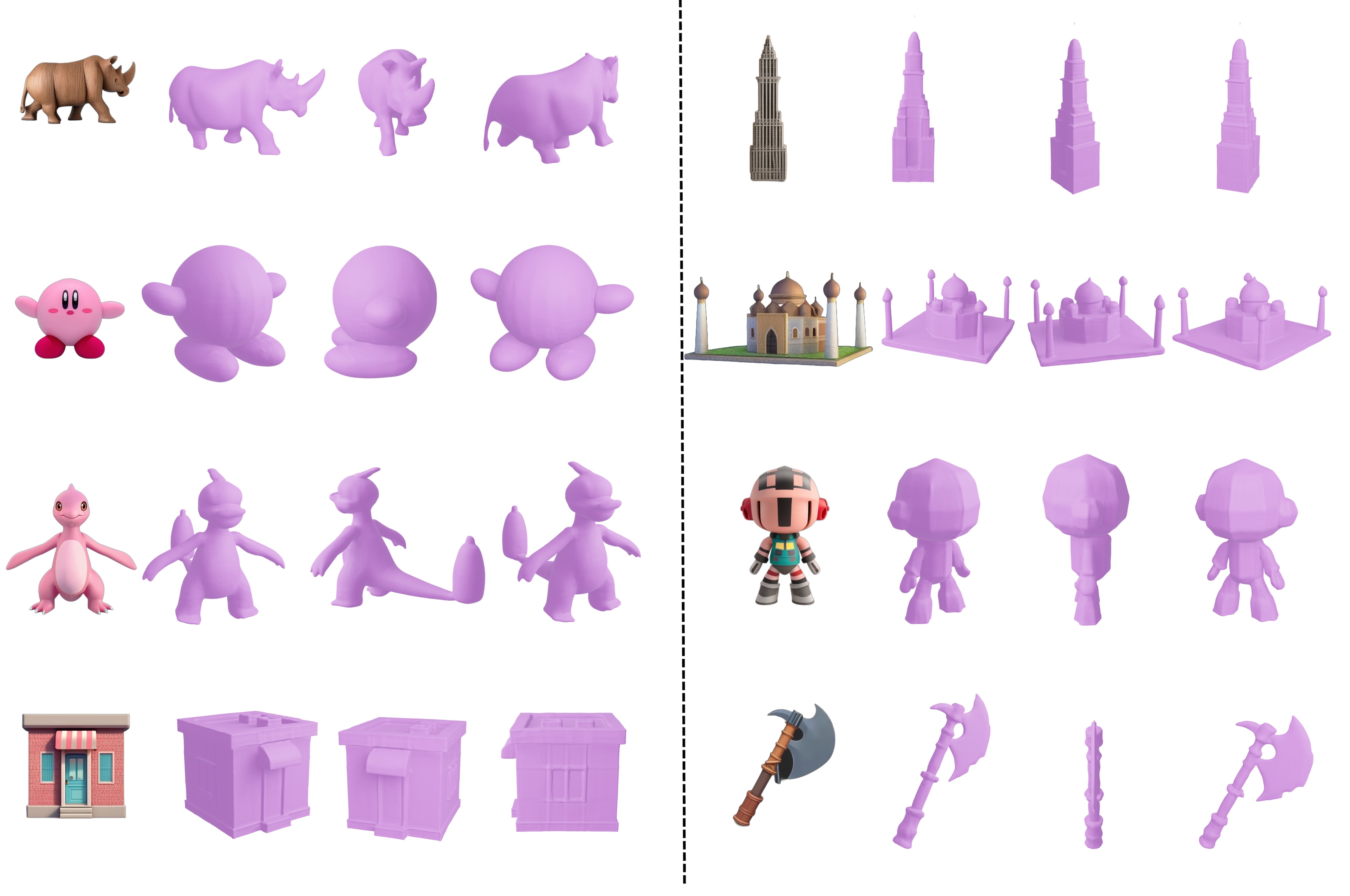} 
\caption{More Visualizations.}
\label{fig: fig_supp_compare_wild}
\end{figure*}

\begin{figure*}[t]
\centering
\includegraphics[width=0.8\textwidth]{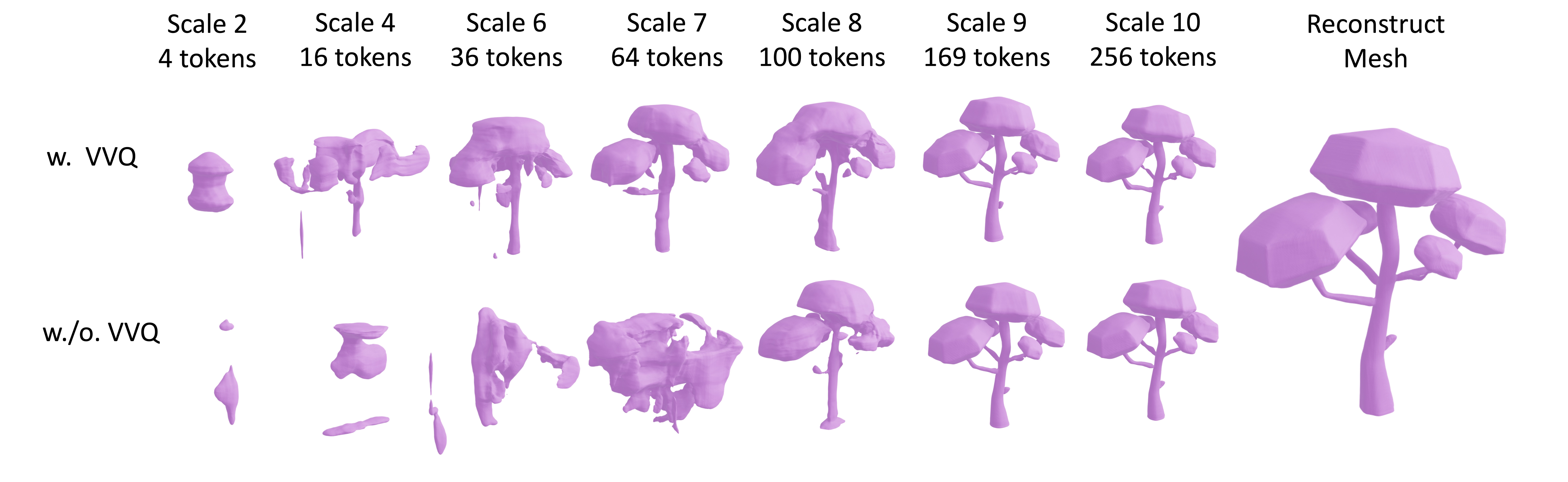} 
\caption{
Visualization of reconstructed mesh from different scales of tokens.
}
\label{fig: multiscale_vis}
\end{figure*}

\begin{figure*}[t]
\centering
\includegraphics[width=0.8\textwidth]{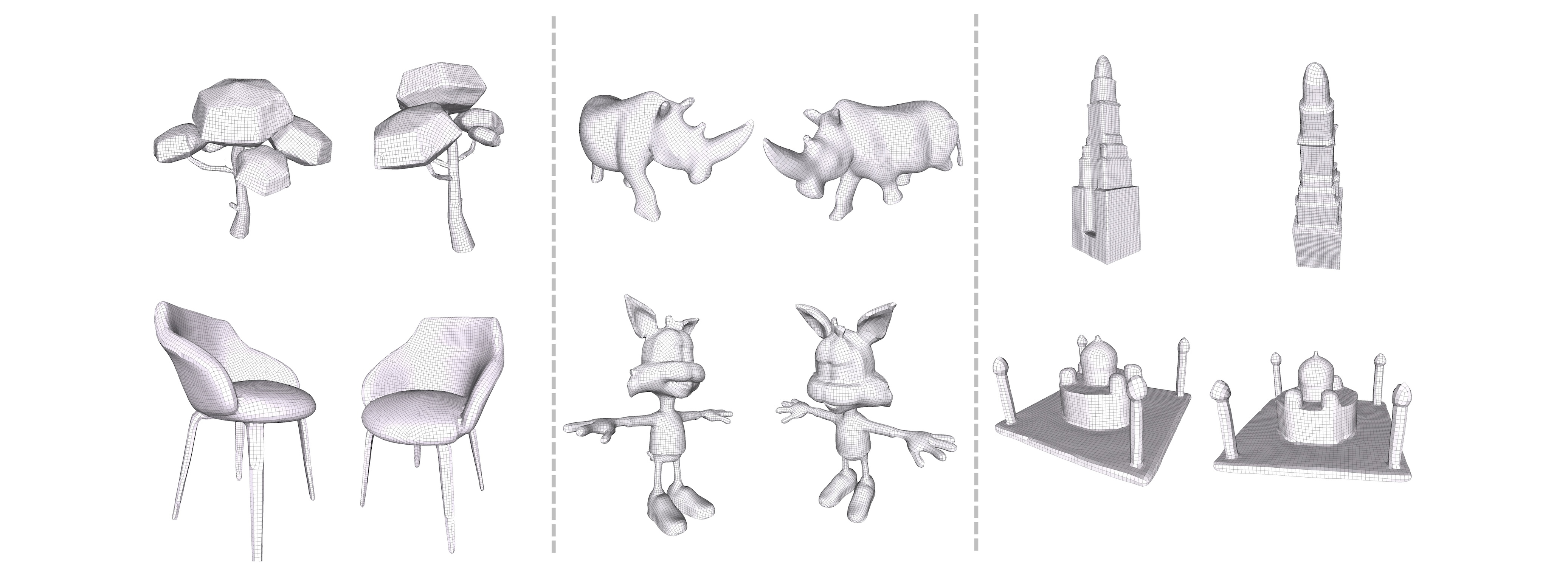} 
\caption{
Quad mesh topologies visualization.
}
\label{fig: multiscale_vis}
\end{figure*}

\begin{figure*}[t]
\centering
\includegraphics[width=1.0\textwidth]{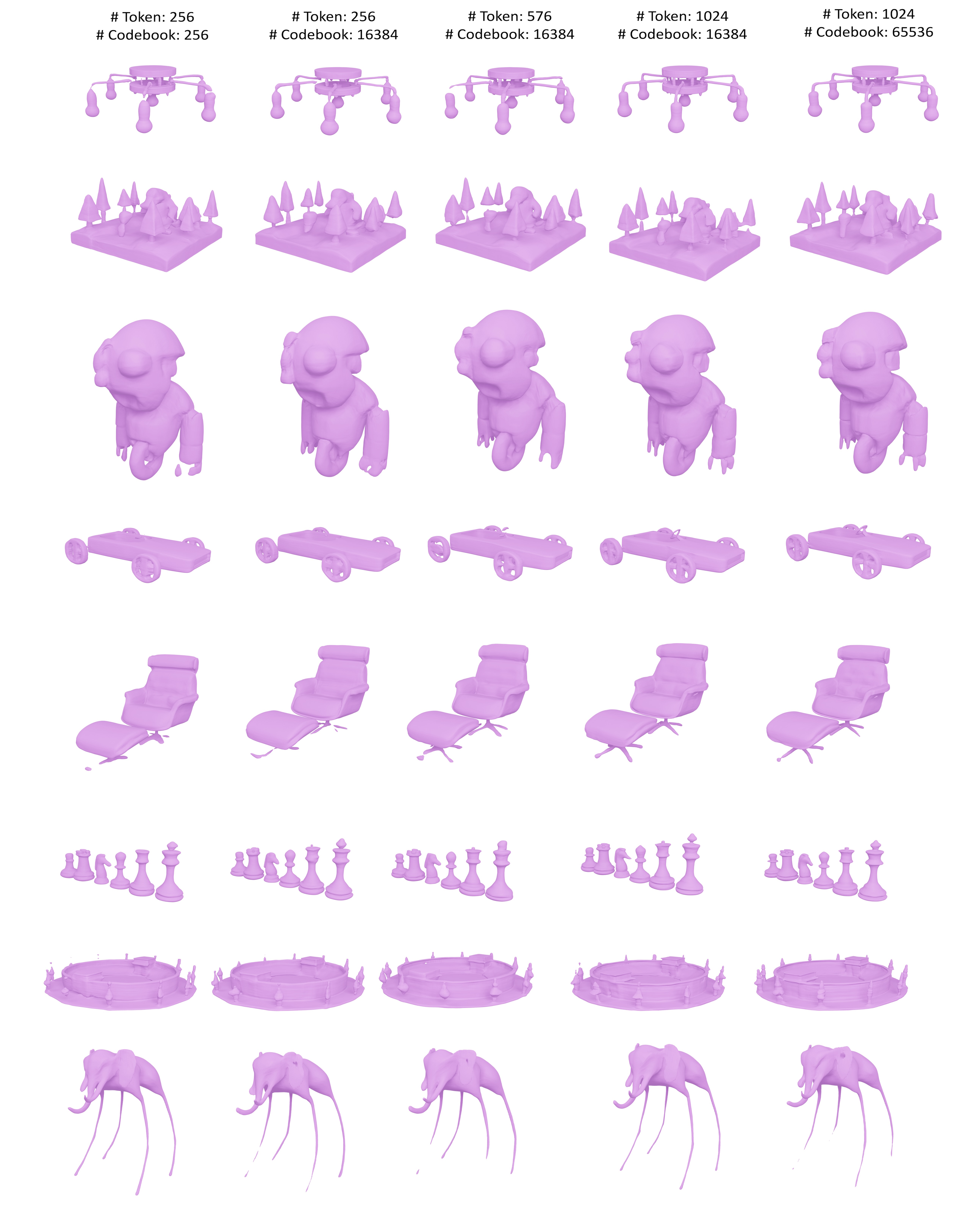} 
\caption{3D reconstruction (surface reconstruction from point clouds) comparison of different VAT variants given different token number and codebook size.}
\label{fig: fig_supp_compare_wild}
\end{figure*}

\begin{figure*}[t]
\centering
\includegraphics[width=1.0\textwidth]{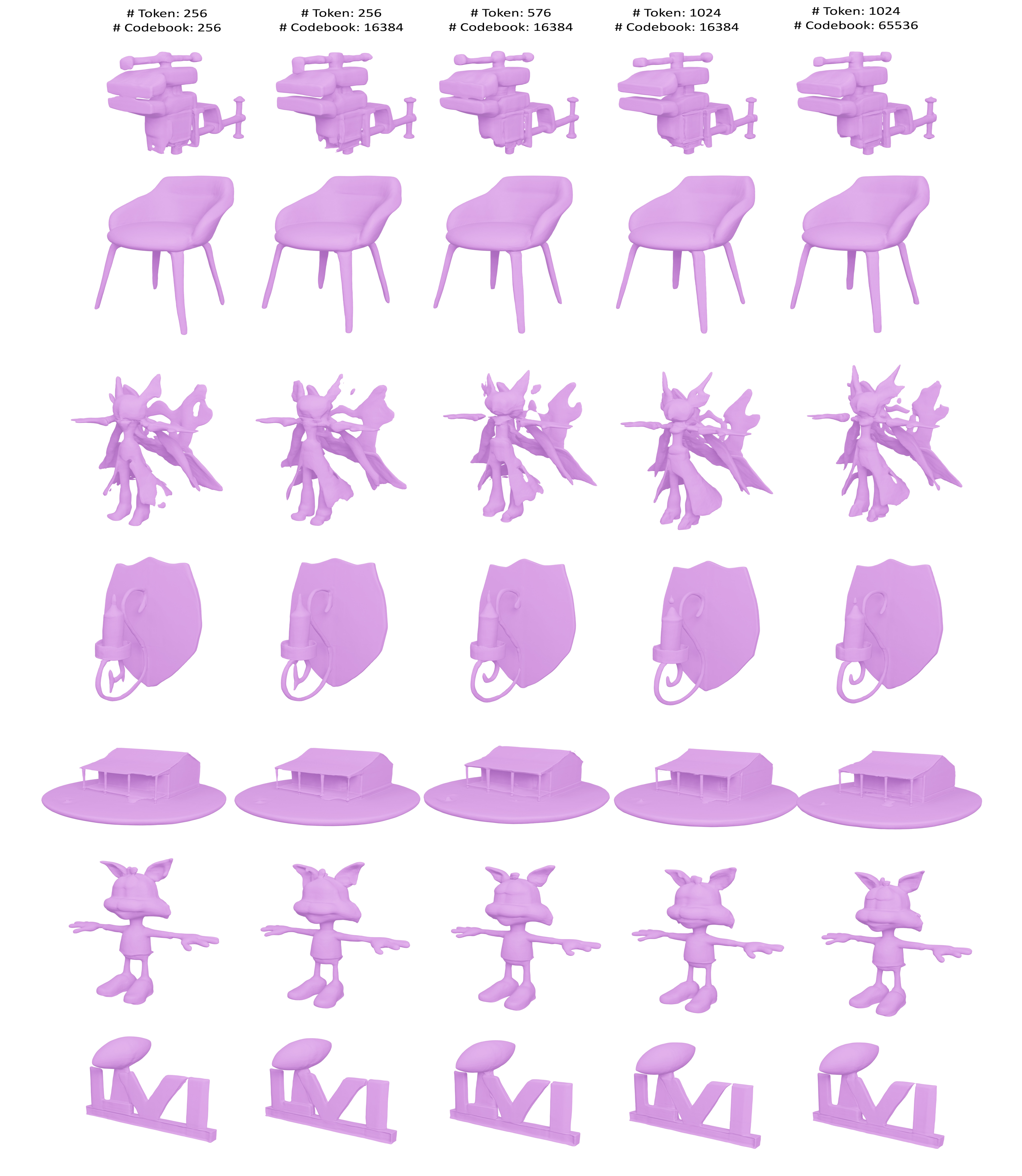} 
\caption{3D reconstruction  (surface reconstruction from point clouds) comparison of different VAT variants given different token number and codebook size.}
\label{fig: fig_supp_compare_wild}
\end{figure*}

\begin{figure*}[t]
\centering
\includegraphics[width=0.9\textwidth]{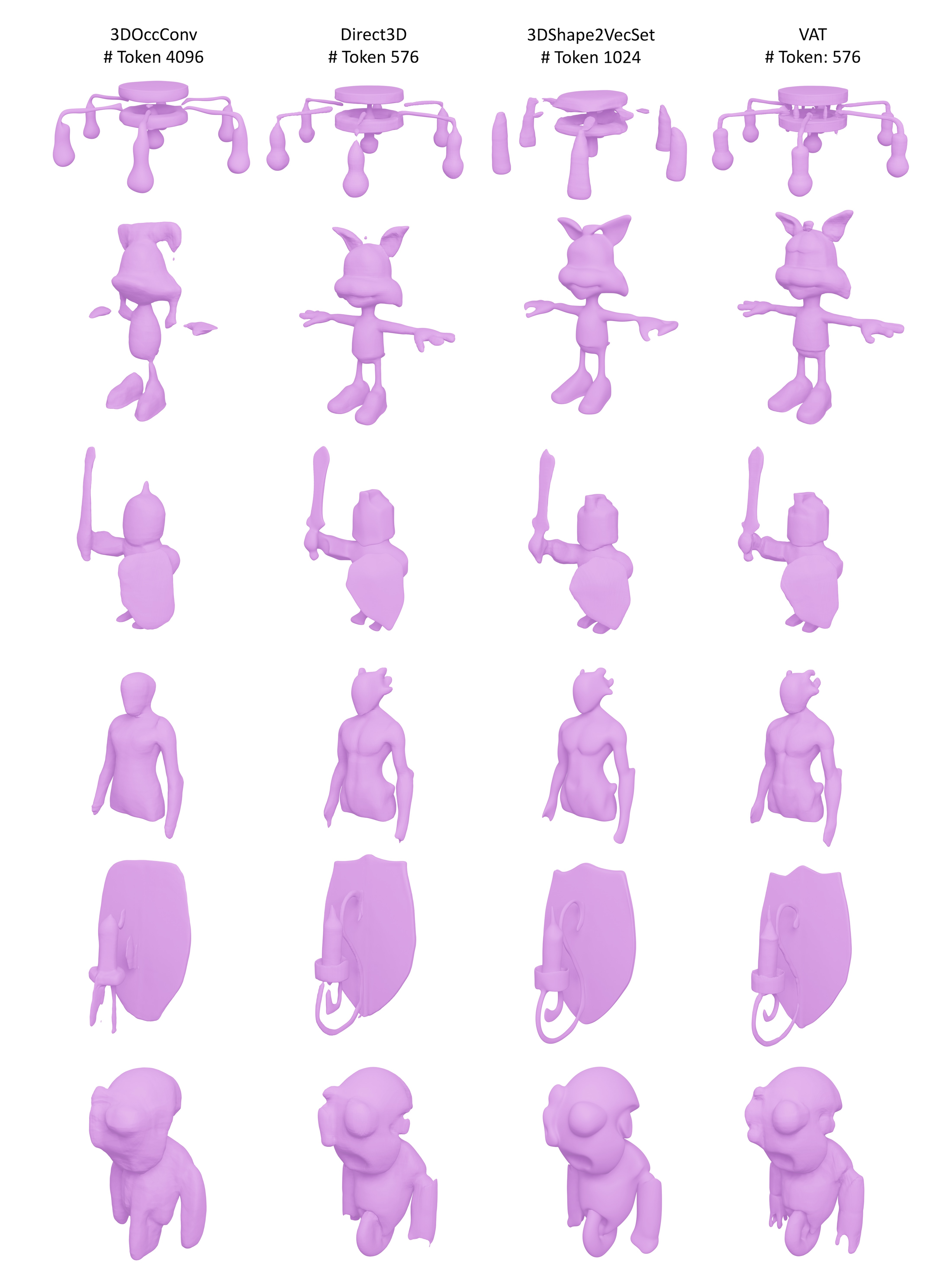} 
\caption{3D reconstruction comparison  (surface reconstruction from point clouds) of different shape autoencoder.}
\label{fig: fig_supp_compare_wild}
\end{figure*}

\begin{figure*}[t]
\centering
\includegraphics[width=0.9\textwidth]{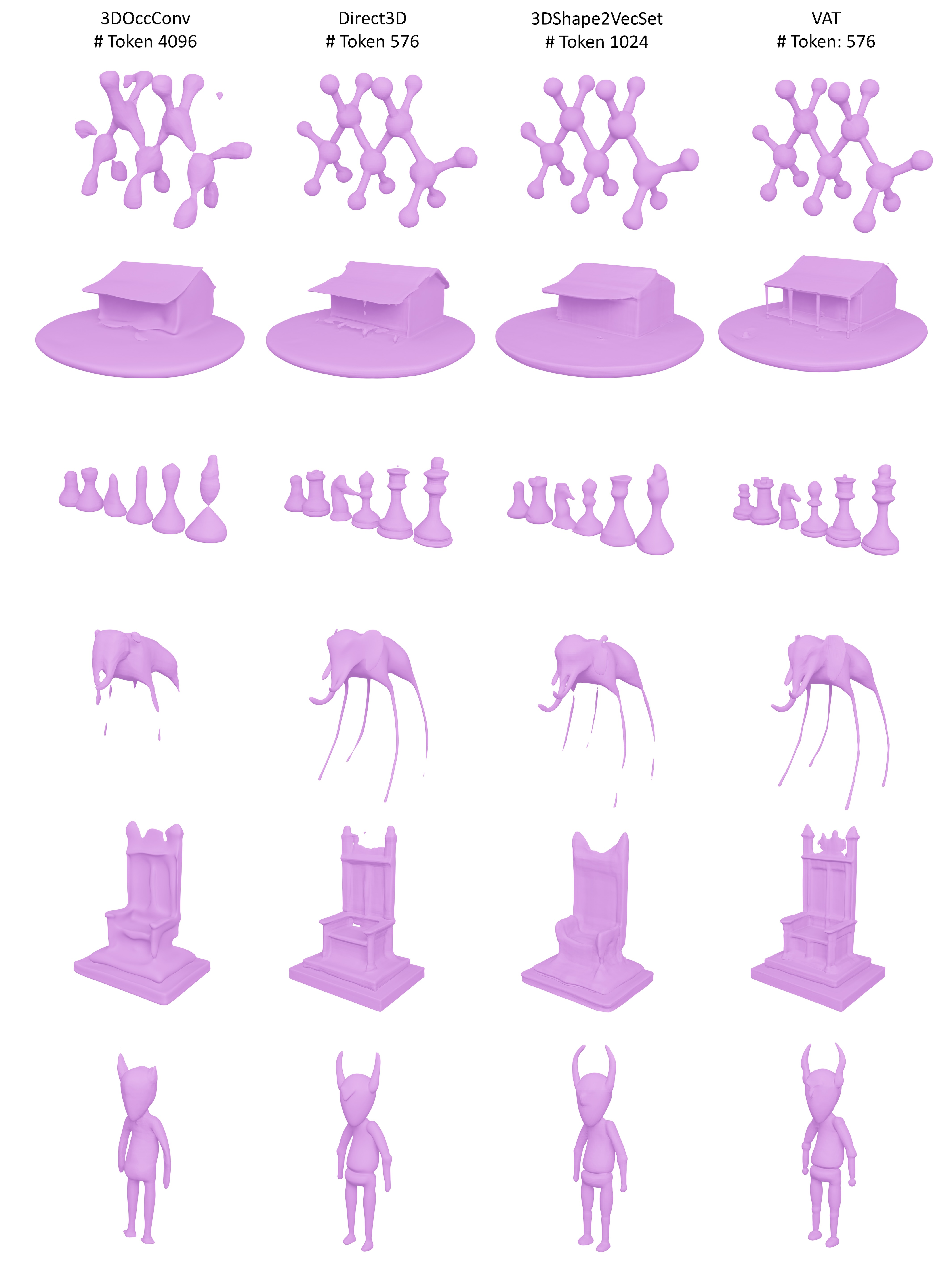} 
\caption{3D reconstruction comparison  (surface reconstruction from point clouds) of different shape autoencoder.}
\label{fig: fig_supp_compare_wild}
\end{figure*}


\end{document}